\documentclass{article}

\usepackage[english]{babel}
\usepackage{graphicx}
\usepackage{framed}
\usepackage[normalem]{ulem}
\usepackage{amsmath}
\usepackage{amsthm}
\usepackage{amssymb}
\usepackage{amsfonts}
\usepackage{enumerate}
\usepackage[dvipsnames]{xcolor}
\usepackage[utf8]{inputenc}
\usepackage[top=1 in,bottom=1in, left=1 in, right=1 in]{geometry}
\usepackage{caption}
\usepackage{subcaption}
\usepackage{float}
\usepackage{verbatim}
\usepackage{braket}
\usepackage{mathtools}
\usepackage{nicefrac}
\usepackage{indentfirst}
\usepackage{url}
\usepackage[colorlinks]{hyperref}
\usepackage{import}
\usepackage{pgf}
\usepackage{lineno}
\usepackage{systeme}
\usepackage{csquotes}
\usepackage{nameref}
\usepackage{authblk}
\usepackage{xfrac}
\usepackage[utf8]{inputenc}
\usepackage[english]{babel}
\usepackage[sorting=none]{biblatex}
\bibliography{references}

\newcommand{\vl}{\underline}
\newcommand{\mc}{\mathcal}

\newcommand{\mr}{\mathrm}
\newcommand{\tl}{\tilde}
\newcommand{\abs}[1]{\left\lvert\!\middle\lvert#1\middle\rvert\!\right\rvert}
\newcommand{\eref}[1]{Eq.~(\ref{#1})}

\newcommand{\aref}[1]{Appendix~\ref{#1}}
\newcommand{\sref}[1]{Section~\ref{#1}} 
\newcommand{\fref}[1]{Fig.~\ref{#1}}

\newcommand{\<}{\left\langle}
\renewcommand{\>}{\right\rangle}

\def\bx{\vl x}
\def\p{P(\bx)}
\def\pplus{P(\bx|y=+)}
\def\pminus{P(\bx|y=-)}
\def\E{\mathbb{E}}

\title{Geometric compression of invariant manifolds in neural nets}
\author[a]{Jonas Paccolat}
\author[a]{Leonardo Petrini}
\author[a]{Mario Geiger}
\author[a]{Kevin Tyloo}
\author[a]{Matthieu Wyart}
\affil[a]{Institute of Physics, \'Ecole Polytechnique F\'ed\'erale de Lausanne, 1015 Lausanne, Switzerland}

\begin{document}

\maketitle

\begin{abstract}

We study how neural networks compress uninformative input space in  models where data lie in $d$ dimensions, but whose label only vary within a linear manifold of dimension $d_\parallel < d$. We show that for a one-hidden layer network initialized with infinitesimal weights (i.e. in the \textit{feature learning} regime) trained with gradient descent, the first layer of weights evolve to become nearly insensitive to the  $d_\perp=d-d_\parallel$  uninformative directions. These are effectively compressed by a factor $\lambda\sim \sqrt{p}$, where $p$ is the size of the training set. 
We quantify the benefit of such a compression on the test error $\epsilon$. 
For large initialization of the weights (the \textit{lazy training} regime), no compression occurs and for regular boundaries separating labels  we find that  $\epsilon \sim p^{-\beta}$, with $\beta_\mr{Lazy} = d / (3d-2)$. Compression improves the learning curves so that $\beta_\mr{Feature} = (2d-1)/(3d-2)$ if $d_\parallel = 1$ and $\beta_\mr{Feature} = (d + \nicefrac{d_\perp}{2})/(3d-2)$ if $d_\parallel > 1$. We test these predictions for a stripe model where boundaries are parallel interfaces ($d_\parallel=1$) as well as for a cylindrical boundary ($d_\parallel=2$).
Next we show that compression shapes the Neural Tangent Kernel (NTK) evolution in time, so that its top eigenvectors become more informative and display a larger projection on the labels. Consequently, kernel learning with the frozen NTK at the end of training outperforms the initial NTK.
We confirm these predictions both for a one-hidden layer FC network trained on the stripe model and for a 16-layers CNN trained on MNIST, for which we also find $\beta_\mr{Feature}>\beta_\mr{Lazy}$. 
The great similarities found in these two cases support that compression is central to the training of MNIST, and puts forward
kernel-PCA on the evolving NTK as a useful diagnostic of compression in deep nets.

\end{abstract}

\section{Introduction and related works}

Deep neural networks are successful at a variety of tasks, yet understanding why they work remains a challenge.
Specifically, the data from which a rule or classes are learnt often lie in high dimension $d$ where the curse of dimensionality is expected. 
Quantitatively, this curse can be expressed on how the test error $\epsilon(p)$ depends on the training set size $p$. If mild assumptions are made on the task (for example regressing a Lipschitz continuous
function), then $\epsilon(p)$ cannot be guaranteed to decay faster than $\epsilon \propto p^{-\beta}$ with an exponent $\beta = \mc O\left(\nicefrac{1}{d}\right)$ \cite{luxburg2004distance}:  learning is essentially impossible. In practice, $\beta$ is found to be much larger and to depend on the task, on the dataset and on the learning algorithm \cite{hestness2017deep,spigler2019asymptotic}, implying that learnable data are highly structured.


Accordingly,  success of neural networks is often attributed to their ability to adapt to the structure of the data, which present many invariances \cite{mallat2016understanding}. For example in the context of classification, some pixels at the edge of the image may be unrelated to the class label. Likewise, smooth deformations of the image may leave the class unchanged. In that view, neural networks correspond to a succession of non-linear and linear operations where directions of neural representation for which the label does not vary are compressed. It is supported by the observations that kernels designed to perform such compression perform well \cite{mallat2016understanding}. 
Yet, there is no quantitative general framework to describe this compression and its effect on the exponent $\beta$. The information bottleneck framework for deep learning \cite{shwartz2017opening} proposes that information is compressed as it propagates deeper in the network. However, information in such a deterministic setting is ill-defined and conclusions can depend qualitatively on details of the architecture or on the estimation of information \cite{saxe2019information}. Still, more robust measures, such as the effective dimension of the neural representation of the data, support that compression occurs in deeper layers \cite{ansuini2019intrinsic,recanatesi2019dimensionality}.

Such a framework should include in which learning regime nets operate. Different regimes have recently been delineated by focusing on the infinite-width limits of neural networks, shown to converge to well-defined learning algorithms \cite{jacot2018neural,rotskoff2018neural,mei2018mean,Du2019}. These are practically useful limits to consider as performance generally improves with width  \cite{neyshabur2017geometry,neyshabur2018towards,bansal2018minnorm,advani2017high,spigler2019jamming}, which simply comes from the fact that convergence to these asymptotic algorithms removes noise stemming from the random initialization of the weights \cite{geiger2020scaling,geiger2019,jacot2020implicit}. Two limits are found, depending on how weights scale with width. In one limit \cite{jacot2018neural}, deep learning becomes equivalent to a kernel method coined Neural Tangent Kernel or NTK. Weights and neuron activities barely change and dimension reduction cannot occur. In the feature learning regime \cite{rotskoff2018neural,mei2018mean}, weights and neuron activities significantly change,  the NTK evolves in time 
\cite{rotskoff2018neural,dou2020training} and compression can in principle occur. Yet understanding this dynamic and its effect on performance remains a challenge. For CNNs the feature learning regime tends to perform better \cite{chizat:hal-01945578,geiger2019,woodworth2020kernel} but it is not so for fully connected nets using vanilla gradient descent on various benchmarks of images \cite{geiger2019}.
This state of affairs calls for simple models of data in which the kernel evolution and its associated compression of invariants can be quantified, together with its effect on performance.

\subsection{Our contribution}

\begin{figure}[H]
    \centering
    \includegraphics[width=\textwidth]{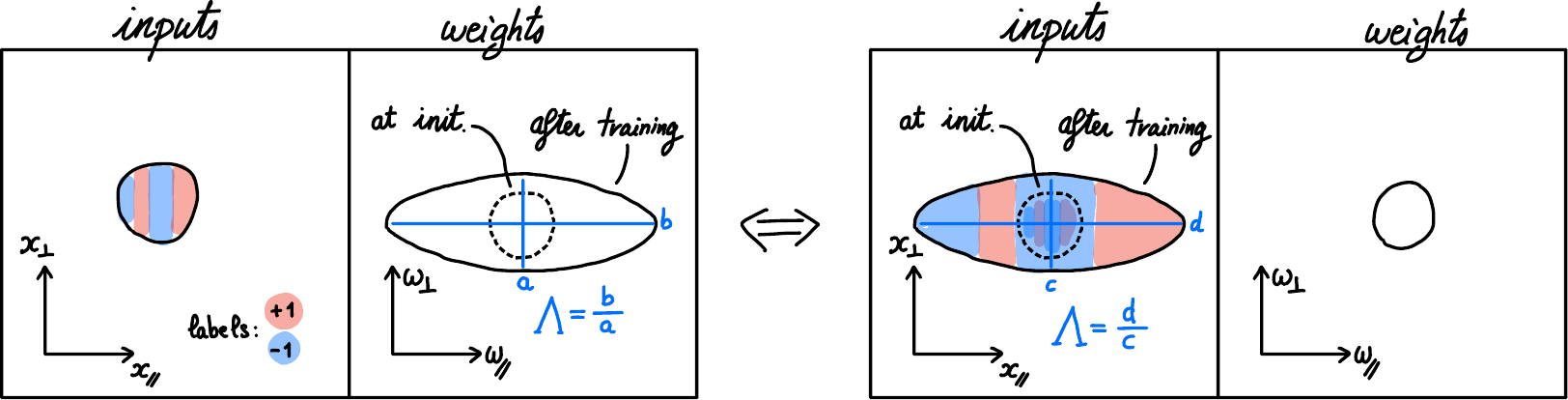}
    \caption{\label{fig:compression_sketch} Illustration of data compression when  the labels do not depend of $d_\perp$ directions in input space, as exemplified in the left panels. During training, the first layer weights  inflate much more in the informative $d_\parallel$ directions. In relative terms, the network thus becomes much less sensitive to the $d_\perp$ uninformative directions. This effect is equivalent to a compression of uninformative directions in  data space, as illustrated on the right panels.}
\end{figure}

Here we consider binary classification and assume that the label does not vary along $d_\perp$ directions of input space. We will first focus on the \textit{stripe model}, arguably the simplest model of invariant yet non linearly-separable data for which  $d_\perp=d-1$, and later show that our results holds for smaller $d_\perp$. Data consists of Gaussian random points $\vl x$ in $d$ dimensions, whose label is a function of a single coordinate $y(\vl x)=y(x_1)$, corresponding to parallel planes separating  labels. In \sref{sec:stripe_model}, we show for the stripe model that:
(i) in the NTK limit, $\beta_\mr{Lazy} = d/(3d-2)$ as we found earlier for isotropic kernels \cite{paccolat2020isotropic}. (ii) In the feature learning regime, if the weights are initialized infinitesimally a geometric compression along invariant directions of magnitude $\lambda\sim \sqrt{p}$ occurs at intermediate times. {This weight compression is equivalent to a spatial compression of the data points as illustrated in \fref{fig:compression_sketch}.} (iii) In the NTK limit if data are compressed by $\lambda$ before learning, performance  closely matches that of the feature learning regime. This observation supports that the main gain of the latter regime is to perform this compression. Assuming that it is the case leads to the prediction $\beta_\mr{Feature} = (2d-1)/(3d-2)$. In \sref{sec:cylinder_model} we generalize this result to the case $d_\perp<d-1$, and argue that for sufficiently regular boundaries separating labels $\beta_\mr{Feature} = (d + \nicefrac{d_\perp}{2}) / (3d - 2)$. We test this prediction when the boundaries separating labels is a cylinder with $d_\perp=1$ and $d=3$.

In  \sref{sec:shaping_kernel}, we argue that the evolution of the NTK is such that at the end of learning: 
(iv) The top eigenvectors of the associated Gram matrix become much more informative on the labels than at initialization. (v) The projection of the labels on these eigenvectors becomes large for the top eigenvectors and small otherwise, supporting that the performance of kernel methods using the NTK improves as it evolves during learning. 
We confirm these predictions empirically in the stripe model. Finally, we show that these points hold true in a multi-layer CNN applied to MNIST data, for which various observables are found to behave very similarly to the stripe model, including the fact that $\beta_\mr{Feature}>\beta_\mr{Lazy}$. These observations support that compression along invariant directions is indeed key to the success of this architecture, and underlines kernel PCA applied to the evolving NTK as a tool to characterize it.

The code used for this article is available online at \url{https://github.com/mariogeiger/feature_lazy/tree/compressing_invariant_manifolds}.

\subsection{Related works}

In the physics literature, $\beta$ has been computed in regression or classification tasks for fixed kernels \cite{gardner1988space,advani2017high,montanari2019generalization,d2020double,dietrich1999statistical}. These results for classification generally consider linearly separable data and apply in the limit $d\rightarrow \infty$ and $p\rightarrow \infty$ with $\alpha=p/d$ fixed. In that limit for specific data it was shown for a regression task that feature learning can outperform the NTK regime \cite{montanari2019generalization}. Here we  consider  classification of non-linearly separable data, and take the limit of large training set size $p$ at fixed dimension $d$ which appears appropriate for common benchmarks \footnote{ MNIST or CIFAR present an effective dimension $d_\mathfrak{M}\in [15,35]$ \cite{spigler2019asymptotic}and $p\approx 6\cdot10^4$. }.

Our work also connects to previous studies on how the anisotropy of the data distribution affects performance \cite{spigler2019asymptotic,goldt2019modelling,paccolat2020isotropic,ghorbani2020neural}. For a large anisotropy, the effective dimension of the data is reduced, improving  kernel methods \cite{spigler2019asymptotic}. The effect of a moderate anisotropy was investigated for kernel classification \cite{paccolat2020isotropic} and regression in neural nets \cite{ghorbani2020neural}. Here we argue that in the presence of invariant, neural nets in the feature learning regime perform a compression equivalent to making the data anisotropic, and to our knowledge produce the first estimates of the training curves rate $\beta$ for both the lazy training and feature learning regime in the limit of large training set size $p$ at fixed dimension $d$.

Guarantees of performance for a one-hidden layer in the feature learning regime exist if some norm (characterizing the magnitude of the weight  representing  the function to be learnt) is finite, and if the dynamics penalizes this norm \cite{rotskoff2018neural,ongie2019function}. In our model that norm is infinite (because there is no margin between labels of different classes). Instead we focus on vanilla gradient descent without special regularization (such regularizations are usually not used in practice). For gradient descent,  with the logistic loss for a one-hidden layer can be shown to correspond to a max-margin classifier in a certain non-Hilbertian space of functions \cite{chizat2020implicit}. Dimension-independent guarantees on performance can be obtained if the data can be separated after projection in a low dimensional space, as occurs in our model. The analysis requires however to go to extremely long times. Here instead we focus on the hinge loss for which the dynamic stops after a reasonable time and we estimate the error and $\beta$ in specific cases instead of providing an upper bound to it.  

On the empirical side, the alignment occurring during learning between the function being learnt and the top eigenvectors of the Gram matrix  was noticed in  \cite{oymak2019generalization} and observed more systematically in \cite{kopitkov2019neural}. Our work offers an explanation for these findings in terms of the compression of invariant directions in data space. 

\section{General considerations on data and dynamics}

\subsection{Linear invariant data}

We consider a binary classification task on data points lying in a $d$-dimensional space whose labels only depend on a linear subspace of dimension $d_\parallel < d$. Without loss of generality, we write the data points as $\vl x = (\vl x_\parallel, \vl x_\perp) \in \mathbb{R}^d$ with $\vl x_\parallel = (x_1, \dots, x_{d_\parallel})$ and $\vl x_\perp = \vl x - \vl x_\parallel$, so that the label function only depends on the $d_\parallel$ first components: $y(\vl x) = y(\vl x_\parallel)$. In this work, we consider data points drawn from the standard normal distribution. In particular, we refer to the points of a training set of size $p$ as $\vl x^\mu \sim \mc N(0, I_d)$, for $\mu=1,\dots,p$.

In \sref{sec:stripe_model}, we shall focus on the simplest case where $d_\parallel = 1$, that we call the stripe model. In \sref{sec:cylinder_model}, we then generalize our findings to higher dimensional tasks and we confirm our results on a ``cylindrical" model with $d_\parallel = 2$. 

\subsection{Learning algorithm}\label{sec:learning_algorithm}

We consider the following fully-connected one-hidden layer neural network of ReLU activation,
\begin{equation}\label{eq:network_function}
f(\vl x) = \frac{1}{h} \sum_{n=1}^h \beta_n \: \sigma\left(\frac{\vl \omega_n \cdot  \vl x}{\sqrt{d}} + b_n\right),
\end{equation}
where $\sigma(x) = \sqrt{2}\max(0, x)$. In our simulations $h=10000$. The trained parameters of the network are $\beta_n$, $\vl \omega_n$ and $b_n$. We use a vanilla gradient descent algorithm with the hinge loss on the predictor function $F(\vl x) = \alpha \left(f(\vl x) - f_0(\vl x)\right)$, where $f_0$ is the network function at initialisation and is not affected by gradient descent. With this trick, the amplitude of the network output is controlled by the scale $\alpha$. Varying it drives the network dynamics from the feature regime (small $\alpha$) to the lazy regime (large $\alpha$) \cite{chizat:hal-01945578}. The dynamical evolution of a generic weight $W \in \{\beta_n, b_n, \vl \omega_n\}_{n=1,\dots, h}$ belonging to the network \eqref{eq:network_function} thus follows the differential equation
\begin{equation}\label{eq:gradient_descent}
    \dot{W} = \frac{1}{p} \sum_{\mu=1}^p \partial_W f(\vl x^\mu) \, y(\vl x_\parallel^\mu) \, l^\prime\left[y(\vl x_\parallel^\mu) F(\vl x^\mu)\right],
\end{equation}
where $l^\prime(x) = \Theta(1 - x)$ is the derivative of the hinge loss. All weights of the network are initialized according to the standard normal distribution. We show in \aref{app:rotation_invariance_2} that the network output is statistically invariant under a rotation of the informative directions. Without loss of generality, we can thus choose the same basis for the data points as for the first layer weights. In particular, we introduce the following notation: $\vl \omega_n = (\vl \omega_{n,\parallel}, \vl \omega_{n,\perp})$.

\subsection{Amplification factor}
The effect of learning is quantified by the compression of the uninformative weights $\vl \omega_{n, \perp}$ with regard to the informative weights $\vl \omega_{n, \parallel}$. Mathematically, the neuron amplification factor $\lambda$ and the global amplification factor $\Lambda$ are defined as 
\begin{equation}
    \lambda_n = \frac{\abs{\vl \omega_{n, \parallel}}_{d_\parallel}}{\abs{\vl \omega_{n, \perp}}_{d_\perp}}
    \hspace{20pt} \text{and} \hspace{20pt}
    \Lambda = \sqrt{\frac{\sum_{n=1}^h \abs{\vl \omega_{n, \parallel}}_{d_\parallel}^2}{\sum_{n=1}^h \abs{\vl \omega_{n, \perp}}_{d_\perp}^2}},
\end{equation}\label{eq:definition_lamnda}
where $d_\perp = d - d_\parallel$ and the $d$-dimensional norm of a vector $\vl v=(v_1, \dots, v_d)$ is defined as $\abs{\vl v}_d^2 = \sum_{i=1}^d v_i^2 / d$, in order to remove the dimensional bias from the ratio.

\subsection{Feature regime vs lazy regime}
Throughout this work, it is assumed that the network width $h$ is sufficiently large for the algorithm to operate in the overparametrized regime \cite{spigler2019jamming, geiger2020scaling}. We define the Neural Tangent Kernel (NTK) $\Theta(\vl x_1, \vl x_2) = \partial_W f(\vl x_1) \cdot \partial_W f(\vl x_2)$, where the scalar product runs over all weights of the network. The gradient descent evolution (\eref{eq:gradient_descent}) on the functional space then reads
\begin{equation}
    \dot f(\vl x) = \frac{1}{p} \sum_{\mu=1}^p \Theta(\vl x, \vl x^\mu) \, y(\vl x_\parallel^\mu) \, l^\prime\left[y(\vl x_\parallel^\mu) F(\vl x^\mu)\right],
\end{equation}
where the NTK can in principle evolve over time.

At initialization, the predictor function is zero. It then grows to fit the training set and doesn't stop until it is at least equal to one on all training points. The smaller the network scale $\alpha$ the more the weights need to evolve.

If $\alpha \gg 1$, the condition $F(\vl x) \sim 1$ can be fulfilled with infinitesimal weight increments $\delta W$, so that the predictor function is linear in $\delta W$. The dynamics thus reduces to a kernel method \cite{jacot2018neural}, meaning that the NTK is frozen to its initial state $\Theta_0$. For an isotropic distribution of the weights, the kernel $\Theta_0$ is isotropic and thus blind to the existence of many invariants in the data to be learned. This regime is coined the lazy regime for finite $h$ or the NTK regime if $h\to\infty$.

\begin{figure}[H]
    \centering
    \includegraphics[width=\textwidth]{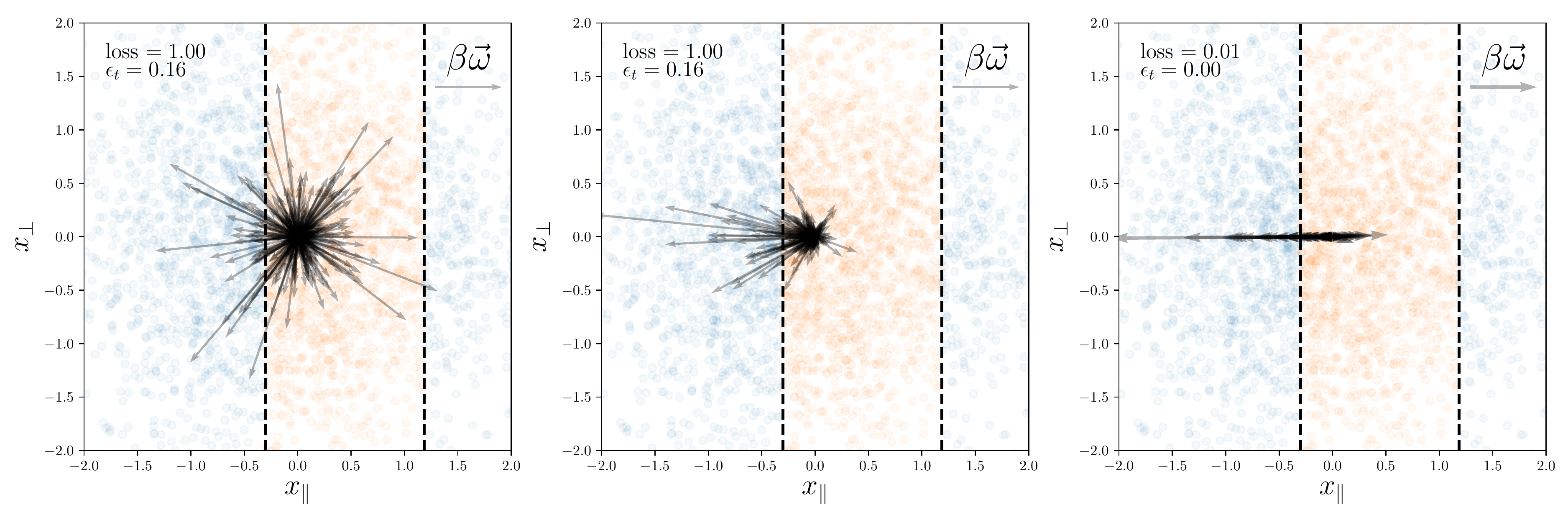}
    \caption{\label{fig:weight_orient_stripe_fromdata} Representation of the weights alignment in the \textbf{single-stripe model}. An instance of the labelled training set is shown in the background. The arrows represent the quantity $\beta \vl \omega$  -- properly rescaled to fit the figure -- for a random subset of neurons. \underline{Left}: At initialization, the weights are distributed isotropically. In the lazy regime, the same distribution persists during the training. \underline{Center} / \underline{Right}: During learning, in the feature regime, the weights tend to align with the direction $\vl e_1$. An animation of the current figure can be found at \href{https://git.io/JJTS9}{git.io/JJTS9}.}
\end{figure}

If $\alpha \ll 1$, the weights of the network need to evolve significantly in order to satisfy the condition $F(\vl x) \sim 1$ \cite{geiger2019}. In that case, the NTK adapts to the data and we shall show that it becomes more and more sensitive to the informative directions. In particular, the first layer weights $\vl \omega_n$ aligns toward the informative linear subspace, as shown in \fref{fig:weight_orient_stripe_fromdata} for the stripe model and in \fref{fig:weight_orient_cylinder_fromdata} for the cylinder model. This regime is coined the feature regime (or sometimes the rich regime) and we study it in the limit $h\to\infty$.

The transition between the two regimes is illustrated in \aref{app:alpha_scan} by learning the stripe model with different values of $\alpha$.

\subsection{Learning timescales}\label{sec:timescales}


We now give a general overview of the network evolution in time. We define the characteristic time $t^\star$ as the time when the predictor function first becomes of order one. Also, we introduce the neuron vector $\vl z = -\sqrt{d} b \vl \omega / \abs{\vl \omega}^2$, which localizes the closest point of the ReLU hyperplane to the origin. We drop the neuron index for simplicity of notation. In the feature regime, we identify three temporal regimes:

\begin{itemize}
    \item Compressing regime: Before $t^\star$, all neuron vectors $\vl z$ converge toward a finite number of fixed points that we generically call $\vl z^\star$ \cite{bousquet2018}. We shall see that the individual weights all diverge exponentially with a time constant $\tau^\star \sim \tau = h \sqrt{\nicefrac{d}{2}}$, which depends on the fixed point. As a consequence, at $t^\star$, the predictor function scales as $\mr{e}^{t^\star/\tau} \alpha \sim 1$. 
    In the mean field limit ($\alpha \to 0$), the characteristic time $t^\star \sim \tau \log(1/\alpha)$ thus diverges and all neurons effectively reach their fixed point. The logarithmic scaling of $t^\star$ is verified numerically in \aref{app:alpha_scan}.
    
    In the limit of infinite training set size ($p\to \infty$), all fixed points are located on the informative subspace, namely $\vl z^\star = (\vl z_\parallel^\star, \vl 0)$. We quantify this compression with the amplification factor $\lambda = \nicefrac{\abs{\vl \omega_\parallel}}{\abs{\vl \omega_\perp}}$ which is divergent in this limit. For finite $p$, the compression is saturated by finite size effects: the data distribution is subject to fluctuations of the order of $\nicefrac{1}{\sqrt{p}}$ compared to its population expectation. The fixed points are thus located at a distance of the order $\mc O\left(\nicefrac{1}{\sqrt{p}}\right)$ perpendicular to the informative subspace. In other words, as we show below the amplification factor saturates at $\lambda \sim \sqrt{p}$.
    \item Fitting regime: After $t^\star$, a finite fraction of the training points satisfy the condition $y^\mu F(\vl x^\mu) > 1$. Because we consider the hinge loss these training points no longer contribute to the network evolution. In particular, they drop out of the sum in \eref{eq:gradient_descent} \cite{giulio2020}. The first points to be excluded are the furthest from the interfaces separating distinct labels. During this process, the fixed points move within the informative manifold such as to better fit the data. Relative fluctuations are still of order $\mc O\left(\nicefrac{1}{\sqrt{p}}\right)$, thus one expects the amplification factor to remain of the same order $\lambda \sim \sqrt{p}$, as we confirm empirically.
    \item Over-fitting regime: When the number of points still remaining in the sum of \eref{eq:gradient_descent} is of the order of one, the sum is dominated by fluctuations and the network overfits the remaining constraints. We check numerically that the previous predictions are not significantly altered during this final regime, which we don't study theoretically.
\end{itemize} 

The neuron compression mechanism scales up to the whole network so that the global amplification factor also saturates with the fluctuations, namely $\Lambda \sim \sqrt{p}$. We expect this scaling to be a general property of linear invariant problems. In the next section, we describe this process in more details for the stripe model.

\section{Stripe model}\label{sec:stripe_model}

We consider the simplest model of linear invariant data, where the label function only depends on $d_\parallel=1$ informative direction, namely $y(\vl x)=y(x_1)$. Layers of $y=+1$ and $y=-1$ regions alternate along the direction $\vl e_1$, separated by parallel planes. In particular, we define the single-stripe model, where the labels are negative if $x_\mr{min} < x_1 < x_\mr{max}$ and positive otherwise. In our numerical simulations, we use this model with the parameters $x_\mr{min} = -0.3$ and $x_\mr{max} = 1.18549$\footnote{The value $x_\mr{max} = \sqrt{2}\,\mr{erf}^{-1}(1 + \mr{erf}(x_\mr{min})) \approx 1.18549$ is chosen so that the two labels are equiprobable.}.

\subsection{Learning curves}\label{sec:learning_curves}
\begin{figure}[ht]
\centering
  \centering
  \hspace*{-0.9cm}
  \includegraphics[scale=.55]{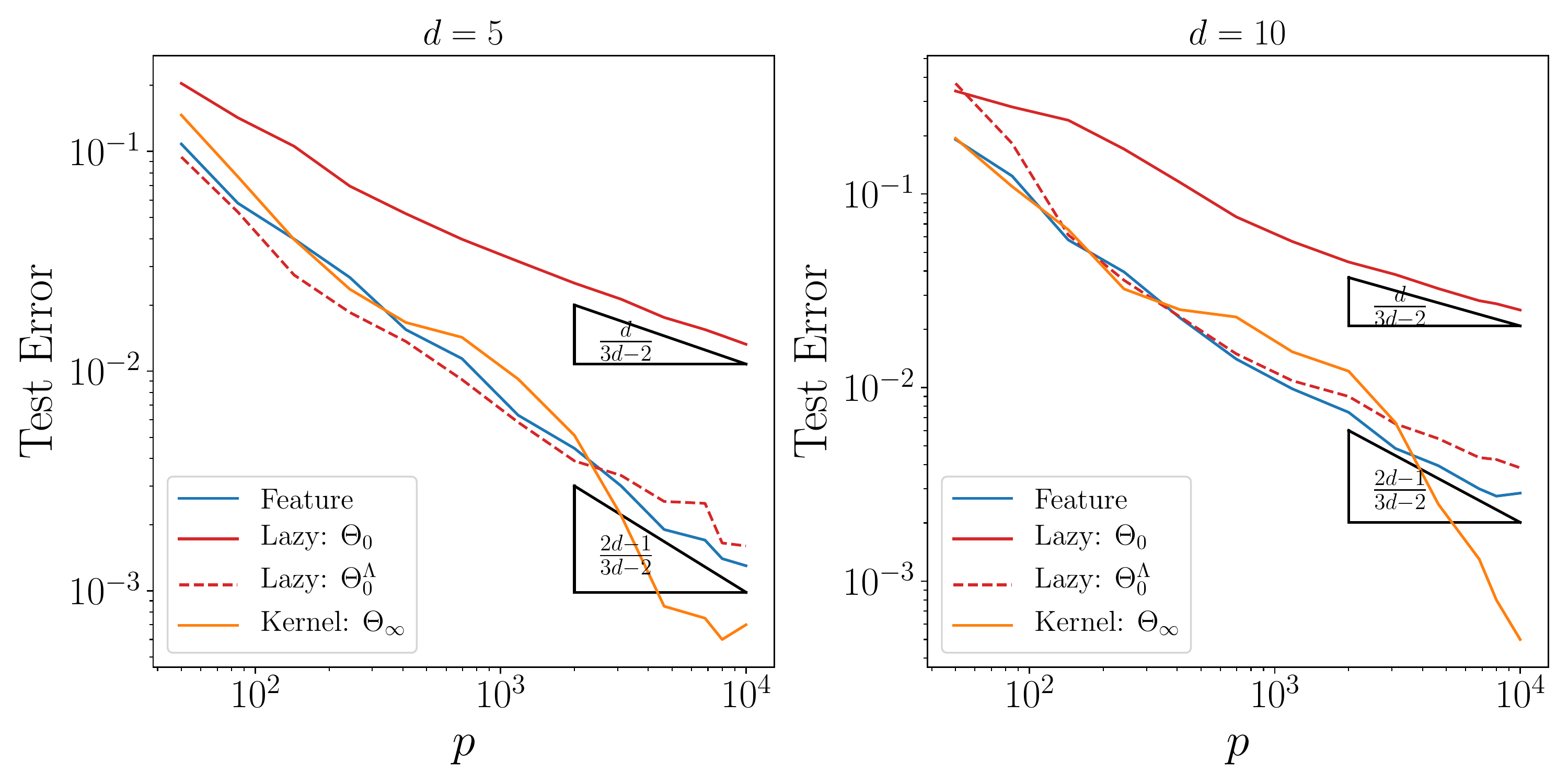}
\caption{\label{fig:learning_curve} Test error vs the training set size $p$ for the \textbf{single-stripe model} in dimension $d=5$ and $d=10$. Two datasets are considered: points drawn from the standard normal distribution in dimension $d$ and its compression, where $\vl x_\perp \to \vl x_\perp / \Lambda^\star$, where $\Lambda^\star \sim \sqrt{p}$ is the global amplification factor at $t
^\star$ (see \sref{sec:amplification_effect} for the definitions). The labels are defined according to the single-stripe model with $x_\mr{min}=-0.3$ and $x_\mr{max} = 1.18549$. The task is learned following the dynamics of \eref{eq:gradient_descent}. In the feature regime (solid blue lines), the network scale is set to $\alpha=10^{-6}$. In the lazy regime, learning is performed with a frozen Gram matrix ($\alpha \to \infty$), computed at initialization for both the original (solid red lines) and compressed (dashed red lines) datasets. The performance of the frozen Gram matrix at the end of feature training is also computed (solid orange lines). 
All results correspond to the median over 20 realizations of both the data distribution and the network initialization. The benchmark triangles represent the expected power laws.}
\end{figure}

We compare the lazy regime and the feature regime, by computing their respective learning curves, namely the test error vs the training set size $p$. \fref{fig:learning_curve} illustrates how the feature regime outperforms the lazy regime, when applied on the single-stripe model.

In the lazy regime, the algorithm reduces to a kernel method and one can rely on \cite{paccolat2020isotropic} to predict the learning curve exponent $\beta$. In that work, it is shown that for an isotropic kernel of bandwidth larger than the distance between nearest neighbours of the training set, the learning curve of the Support Vector Classifier (SVC) algorithm applied to the stripe model in dimension $d$ scales as $\epsilon \sim p^{-\beta}$, with $\beta = (d - 1 + \xi)/(3d-3 + \xi)$, where $\xi \in (0, 2)$ is an exponent characterizing the kernel cusp at the origin. The NTK is isotropic on data lying on the sphere, has a bandwidth of order $\mc O(1)$ and its cusp is similar to the one of a Laplace kernel, namely $\xi = 1$. Hence, as the SVC algorithm minimizes the hinge loss, the learning curve of the lazy regime is expected to have an exponent $\beta_\mr{Lazy} = d/(3d-2)$. This prediction is tested on \fref{fig:learning_curve}.

In the same work, it is shown that if the uninformative directions of the data are compressed by a factor $\Lambda$, namely $\vl x_\perp \to \vl x_\perp / \Lambda$, the test error is improved by a factor $\Lambda^{-\frac{2(d-1)}{3d-2}}$ for $\xi = 1$. In the next section, we shall argue that, in the feature regime, the perpendicular weights $\vl \omega_{n, \perp}$ are suppressed by a factor $\sqrt{p}$ compared to the informative weights $\omega_{n, 1}$ as their growth is governed by fluctuations of the data. Such a weight compression acts similarly as a data compression with $\Lambda \sim \sqrt{p}$ as depicted on \fref{fig:compression_sketch}. Assuming that the main effect of feature learning is this compression, we expect the learning curve exponent of the feature regime to be $\beta_\mr{Feature} = (2d-1)/(3d-2)$. This scaling is again consistent with the numerical results of \fref{fig:learning_curve}.

\subsection{Amplification effect}\label{sec:amplification_effect}

In this section we show that when learning the stripe model in the feature regime, the first layer weights align along the informative direction $\vl e_1$. In particular, we show that the ratio between the informative (or parallel) weights $\omega_{n, 1}$ and the uninformative (or perpendicular) weights $\vl \omega_{n, \perp}$ scales as $\Lambda\sim \sqrt{p}$. This section being more technical can be skipped at first reading. For the interested reader, details are given in \aref{app:stripe_analytics}.

\subsubsection{Neuronal dynamics}

We first consider the dynamics of a single generic neuron, whose dynamics is obtained from \eref{eq:gradient_descent}:
\begin{align}\label{eq:neuron_dynamics1}
\dot{\vl \omega} &= \frac{1}{hp} \, \sum_{\mu=1}^p \, \sigma^\prime\left[\frac{\vl \omega \cdot \vl x^{\mu}}{\sqrt{d}} + b \right] \, l^\prime\left[y^{\mu} F(\vl x^\mu)\right] \, \beta \, \frac{\vl x^\mu}{\sqrt{d}} \, y^{\mu} \nonumber \\
\dot{b} &= \frac{1}{hp} \, \sum_{\mu=1}^p \, \sigma^\prime\left[\frac{\vl \omega \cdot \vl x^{\mu}}{\sqrt{d}} + b \right] \, l^\prime\left[y^{\mu} F(\vl x^\mu)\right] \, \beta \, y^{\mu} \,, \\
\dot{\beta} &= \frac{1}{hp} \sum_{\mu=1}^p \, \sigma\left[\frac{\vl \omega \cdot \vl x^{\mu}}{\sqrt{d}} + b \right] \, l^\prime\left[y^{\mu} F(\vl x^\mu)\right] \, y^{\mu} \, \nonumber
\end{align}
where the neuron index $n$ is dropped. Because the ReLU activation is homogeneous, $\sigma(x) = x \sigma^\prime(x)$, the equality $\beta \dot\beta - \vl \omega \cdot \dot{\vl \omega} - b \dot b = 0$ holds during the whole evolution. Following the discussion of \sref{sec:timescales}, we now solve the above system in the limit $\alpha \to 0$, so that $t^\star \sim \tau \log (1/\alpha) \to \infty$. In the numerical experiments, we choose the network scale $\alpha=10^{-6}$ and define $t^\star$ as the time when 10\% of the training set satisfies the condition $y^\mu F(\vl x^\mu) > 1$.

\paragraph{Compressing regime}\label{sec:compressing_regime}
As long as $t \ll t^\star$, the quantity $l^\prime\left[y^\mu F(\vl x^\mu)\right]=1$, $\forall \mu$, so that the system \eqref{eq:neuron_dynamics1} only depends on the weights associated to the considered neuron. Each neuron thus evolves independently and only differs from the other neurons by its initial conditions.

We first consider the limit $p\to\infty$ and neglect the finite size effects. Applying the central-limit theorem, we carry out the integration over the perpendicular space in \aref{app:computation_averages}. Defining the neuron amplification factor $\lambda = \omega_1 / \omega_\perp$, where $\omega_\perp = \abs{\vl \omega_\perp}$, the neuronal dynamics \eqref{eq:neuron_dynamics1} becomes
\begin{align}\label{eq:neuron_dynamics2}
    \dot{\omega}_1 &= \frac{\beta}{\tau} \, \< y(x_1) \, x_1 \, g_\lambda(x_1 - \zeta_1)\>_{x_1} \, + \, \mc O\left(p^{-\nicefrac{1}{2}}\right) \nonumber \\
    \dot{\vl \omega}_\perp &= \frac{\beta}{\tau} \, \frac{\mr{e}^{-\nicefrac{db^2}{2\omega^2}}}{\sqrt{2\pi}} \, \frac{\vl \omega_\perp}{\omega} \< y\left(\frac{\omega_\perp}{\omega}x_1 + \frac{\omega_1^2}{\omega^2}\zeta_1\right) \>_{x_1} \, + \, \mc O\left(p^{-\nicefrac{1}{2}}\right) , \\
    \dot{b} &= \frac{\sqrt{d}\beta}{\tau} \, \< y(x_1) \, g_\lambda(x_1 - \zeta_1)\>_{x_1} \, + \, \mc O\left(p^{-\nicefrac{1}{2}}\right)  \nonumber
\end{align}
where $\tau = h\sqrt{\nicefrac{d}{2}}$, $\omega = \abs{\vl \omega}$ and $\zeta_1 = -\sqrt{d} b / \omega_1$ is the intercept of the ReLU hyperplane with the $\vl e_1$ axis, while $g_\lambda(x) = \frac{1}{2} \left(1 + \mr{erf}(\lambda x / \sqrt{2})\right)$. The notation $\<\boldsymbol{\cdot} \>_{x_1}$ refers to the expectation over the Gaussian variable $x_1$.

\begin{figure}[H]
\centering
\includegraphics[width=\linewidth]{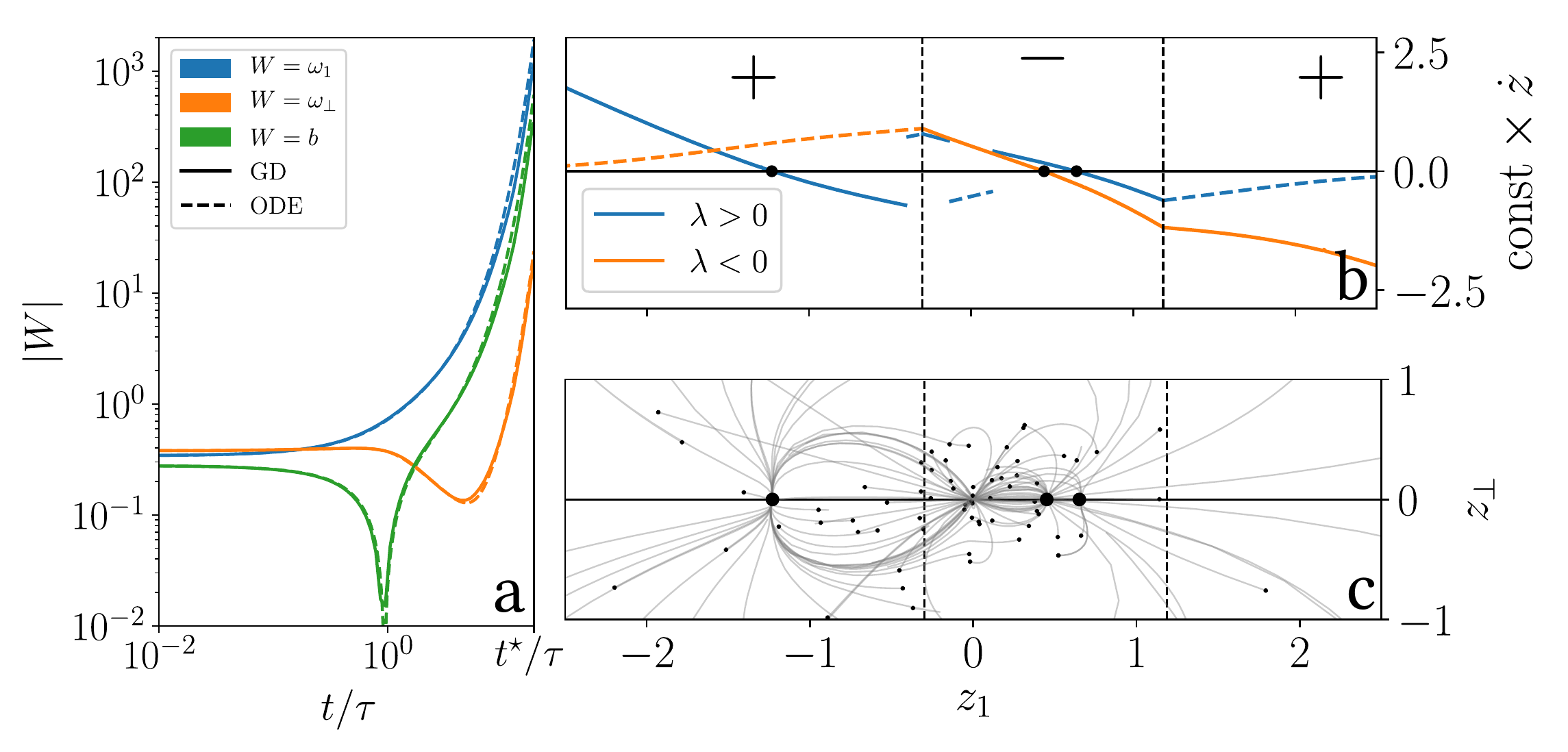}
\caption{\label{fig:zprime} Numerical analysis of the \textbf{single-stripe model} with $x_\mr{min}=-0.3$ and $x_\mr{max}=1.18549$ in dimension $d=2$ with a training set of size $p=10000$. The location of the two interfaces is illustrated by the vertical dashed lines. \textbf{(a)} Temporal evolution of the weights of a randomly chosen neuron. The solid lines illustrate the considered neuron dynamics in the neural network, while the dashed lines correspond to the numerical solutions of the ODE \eref{eq:neuron_dynamics2} and \eref{eq:perpendicular_weight_dynamics} obtained for the same initial conditions. The random variables $\vl N (\vl z^\star)$ is computed numerically. The curves are truncated at the time $t^\star$. \textbf{(b)} Function defining the location of the fixed points along $\zeta_1$ in the limit $\lambda \to \infty$. The two scenarios $\lambda > 0$ and $\lambda < 0$ are shown. The unstable regions, where the limit $\lambda \to \infty$ is inconsistent are represented with dashed lines. \textbf{(c)} Selection of neuronal trajectories in the $\vl z$-plane for $t<t^\star$. The small black dots mark the location of the initial conditions, while the large black dots lie on the predicted location of the three attractors of the compressing regime.}
\end{figure}

We recall the definition of the neuron vector $\vl z = - \sqrt{d} b \vl \omega / \omega^2$. In \cite{bousquet2018}, the authors show that the first layer weights of a one-hidden layer network of ReLU activation tend to align along a finite number of directions depending only on the dataset. Relying on the symmetries of the model, we seek solutions on the informative axis. We thus make the hypothesis that the fixed points are of the form $\vl z^\star = (z^\star, \vl 0)$, where $z^\star = z_1^\star = \zeta_1^\star$, which is equivalent to assuming that the amplification factor associated to such fixed points is diverging. In this limit, the system \eqref{eq:neuron_dynamics2} simplifies: the expectation values only depend on the parameter $\zeta_1$ and the sign of $\lambda$. We respectively call them $C^\pm_1(\zeta_1)$, $C^\pm_\perp(\zeta_1)$ and $C^\pm_b(\zeta_1)$. As a consequence, the dynamics of $\zeta_1$,
\begin{equation}
    \dot \zeta_1 \xrightarrow{\lambda\to\pm\infty} -\frac{1}{\tau}\frac{\beta}{\omega_1} [d \, C^\pm_b(\zeta_1) + \zeta_1 \, C^\pm_1(\zeta_1)],
\end{equation}
yields the location of the fixed points as they lie where the above bracket vanishes. For the fixed points to be stable along the $\vl e_1$ axis, the second derivative of $\zeta_1$ needs to be negative. On a given fixed point $z^\star$ the expectation values $C^\pm_1(z^\star)$, $C^\pm_\perp(z^\star)$ and $C^\pm_b(z^\star)$ are constant and it is straight-forward to see that $\omega_1$, $b$ and $\beta$ all diverge exponentially with a time constant $\tau^\star \sim \tau$ given in \aref{app:infinite_p}. Finally, we verify in \aref{app:infinite_p} that the perpendicular weights do not diverge as fast as $\omega_1$ as long as $\lambda C^\pm_1(z^\star) y(z^\star) < 0$ or $\sqrt{2\pi} |C^\pm_1(z^\star)| - \mr{e}^{-z^{\star 2}/2} > 0$. Under these conditions, the amplification factor $\lambda$ thus diverges exponentially in time which justifies our initial hypothesis. We checked numerically that these conditions indeed hold for the considered models. The panel $\textbf{b}$ of \fref{fig:zprime} illustrates $\dot \zeta$ for the single-stripe model in $d=2$.

We now consider the finite $p$ corrections to a given fixed point $\vl z^\star$ and show that the amplification factor saturates at $\lambda^\star \sim \sqrt{p}$. The finite $p$ effects lead to an additional fluctuation term in each equation of the system \eqref{eq:neuron_dynamics1}. This correction is negligible for the dynamics of $\omega_1$, $b$ and $\beta$, however for the perpendicular weights it yields 
\begin{equation}\label{eq:perpendicular_weight_dynamics}
    \dot{\vl \omega}_\perp = \frac{\beta}{\tau} \, \left[ \frac{\mr{e}^{-\nicefrac{z^{\star 2}}{2}}}{\sqrt{2\pi}} \, \frac{\vl \omega_\perp}{\omega} C^\pm_\perp(z^\star) + \frac{\vl N(z^\star)}{\sqrt{p}} D^\pm_\perp(z^\star) \right],
\end{equation}
where $D^\pm_\perp(z^\star) = \<\Theta(\pm(x_1 - z^\star))\>_{x_1}$ and $\vl N(\vl z^\star)$ is a vector of random variables of variance one (see \aref{app:finite_p}). The first term in the above bracket is proportional to $\nicefrac{1}{\lambda}$ and thus vanishes exponentially with time until it is of the order of the second term, namely $\mc O\left(\nicefrac{1}{\sqrt{p}}\right)$. We call $\tau_\perp$ the time when this crossover occurs. After $\tau_\perp$, $\dot{\vl \omega}_\perp$ is merely proportional to $\beta / \sqrt{p}$. Therefore, the perpendicular weights follow the same exponential growth as the other weights up to a $\mc O\left(\nicefrac{1}{\sqrt{p}}\right)$ prefactor and the amplification factor converges to a finite value $\lambda^\star$ that scales as 
\begin{equation}
    \lambda^\star \sim \sqrt{p}.
\end{equation}

We test numerically that all neurons converge to one of the above described fixed points by considering the single-stripe model. The panel \textbf{c} of \fref{fig:zprime} illustrates the trajectories of a random selection of neurons while training the network \eref{eq:network_function} until $t^\star$. Note that some neurons may not have yet reached a fixed point for two reasons. First, because $p$ is finite, a neuron initial position may lie too far from the training set domain. If no training point lies within the positive side of its associated ReLU hyperplane, it won't feel any gradient and will thus remain static. Second, the simulation is run with a finite network scale ($\alpha = 10^{-6}$), implying that the time $t^\star \sim  \tau \log(1/\alpha)$ is also finite. Hence, some neurons may not have reached their asymptotic regime at $t^\star$\footnote{Because $\lambda$ initially grows exponentially, the definition of the perpendicular timescale yields $\tau_\perp \sim \tau^\star \log p$. For the amplification factor to reach its plateau $\lambda^\star \sim \sqrt{p}$ during the compressing regime, it is essential that $\tau_\perp < t^\star \sim \tau \log(1/\alpha)$. Hence the larger the training set size, the smaller $\alpha$ needs to be.}.

\paragraph{Fitting regime}
After $t^\star$, the loss derivative is zero on a finite fraction of the training set. As discussed in \sref{sec:timescales}, these training points no longer contribute to the network dynamics. This long time evolution is beyond the scope of this work, but could be solved numerically in the limit $p\to\infty$ following the work of \cite{giulio2020}. It requires to compute the network function at each step in order to decide which training points still contribute to the dynamics.

\begin{figure}[H]
\centering
\includegraphics[width=1\linewidth]{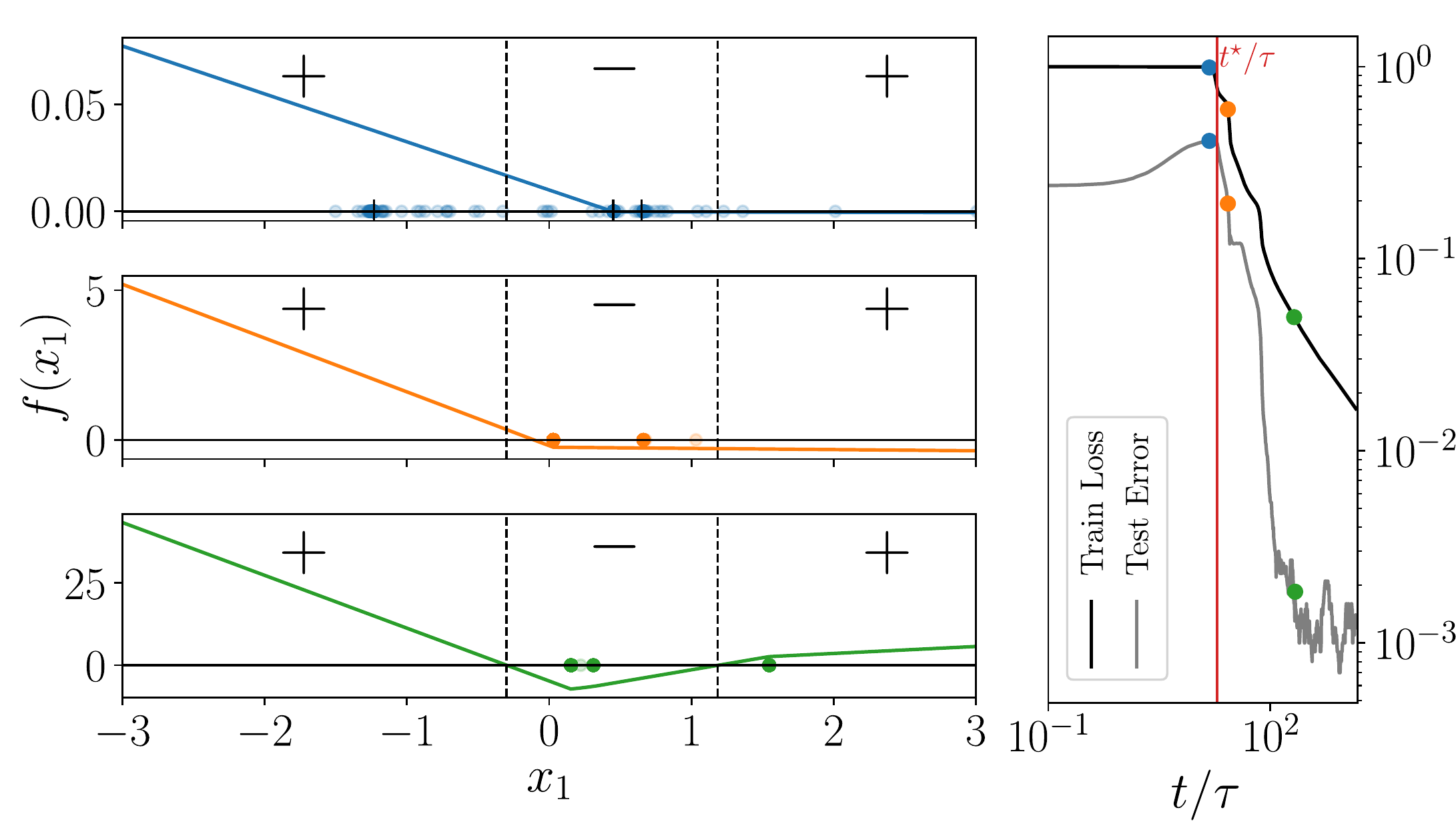}
\caption{\label{fig:function_evolution} Evolution of the network while fitting the \textbf{single-stripe model} in $d=2$ with a training set of size $p=10000$. \underline{Left}: Neural network decision function along the informative direction $\vl e_1$ at three different times. The $\zeta_1$ variable of each neuron is represented on the $x$-axis by colored dots. The darker the region, the larger the point density. On the top plot, the location of the predicted fixed points is marked by black crosses. The location of the two interfaces is illustrated by the vertical dashed lines. \underline{Right}: Train loss and test error vs time. The three times considered on the left plot are indicated with the same color code. The characteristic time $t^\star$ is represented by the vertical red line.}
\end{figure}

In this regime, the neurons are still sparsely distributed on the same number of fixed points \cite{bousquet2018} as in the previous regime. The location of the fixed points is however changing to fit the stripe. This process is shown on \fref{fig:function_evolution} for the stripe model in $d=2$. Concerning the amplification factor, the $\sqrt{p}$ suppression of $\dot{\vl \omega}_\perp$ compared to $\dot{\omega}_1$ remains true until the effective number of training points contributing to the dynamics becomes of order $\mc O(1)$, as shown on the top panel of \fref{fig:amplification_factor}.

\subsubsection{Global amplification factor}\label{sec:global_amplification_factor}

\begin{figure}[H]
\centering
\includegraphics[width=.75\linewidth]{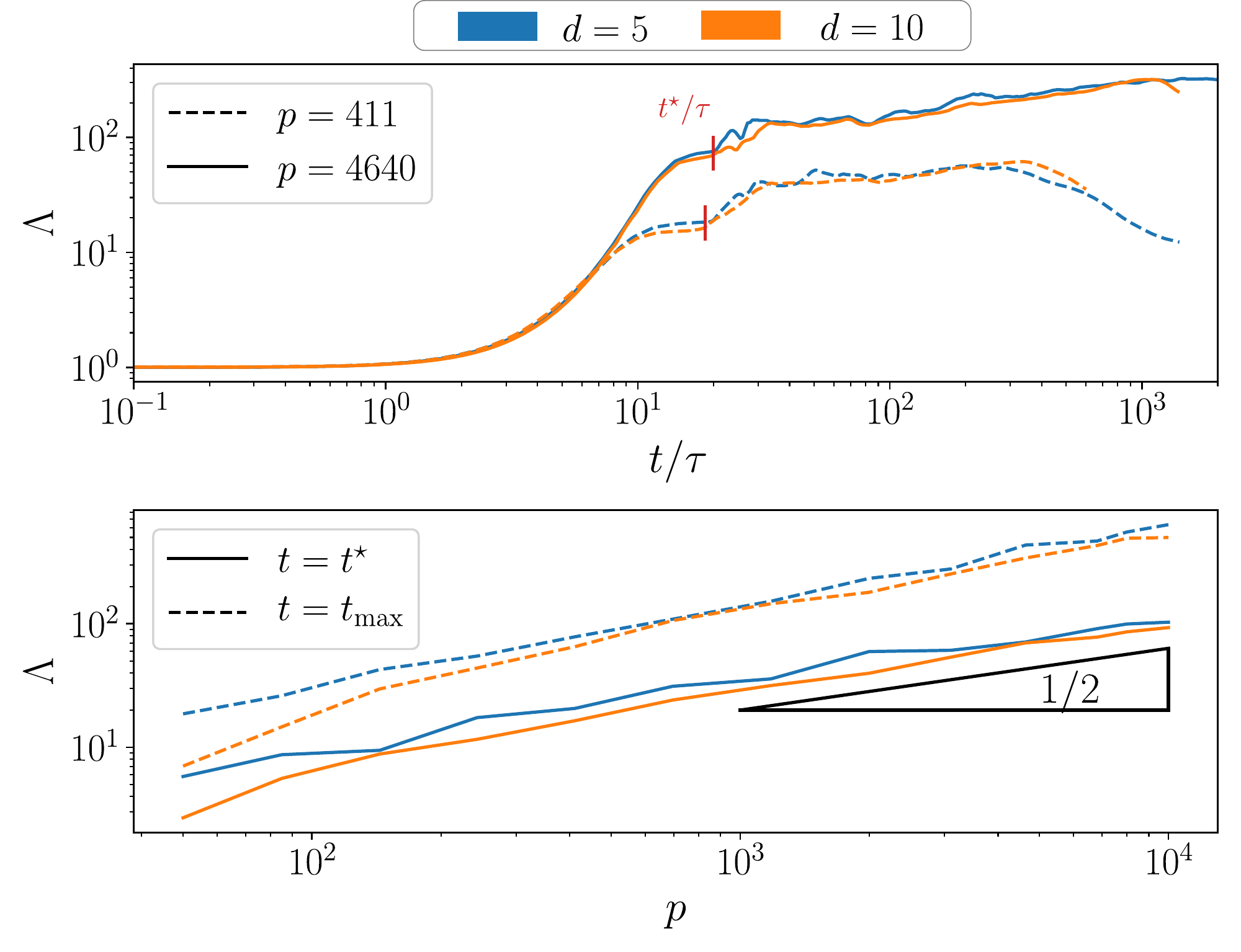}
\caption{\label{fig:amplification_factor} \underline{Top}: Temporal evolution of the global amplification factor while learning the \textbf{single-stripe model}. Two dimensions $d=5$ and $d=10$ and two training set sizes $p=411$ and $p=4640$ are illustrated. The curves are averaged over 20 realizations of both the data distribution and the network initialization. The red vertical ticks mark the averaged critical time of the associated setup. \underline{Bottom}: Global amplification factor vs the size of the training set $p$ for the \textbf{single-stripe model} in dimensions $d=5$ and $d=10$. Both the amplification factor $\Lambda^\star$ computed at $t^\star$ and the maximal amplification factor $\Lambda_\mr{max}$ are displayed. The curves correspond to the median over 20 realizations of both the data distribution and the network initialization. The benchmark triangle of slope $\nicefrac{1}{2}$ confirms our scaling predictions for $\Lambda^\star$ and $\Lambda_\mr{max}$.}
\end{figure}

In the previous discussion, we defined an amplification factor $\lambda = \omega_1 / \omega_\perp$ for each neuron of the network. Following the definition \eref{eq:definition_lamnda} we now consider the global amplification factor $\Lambda$ averaged over all neurons, namely
\begin{equation}
    \Lambda^2 = (d-1) \frac{\sum_{n=1}^h \omega_{n, 1}^2}{\sum_{n=1}^h \omega_{n, \perp}^2}.
\end{equation}
This definition compares the largest parallel weights to the largest perpendicular weights. The prefactor guarantees that $\Lambda(t=0) = 1$. The top panel of \fref{fig:amplification_factor} shows the exponential growth of $\Lambda$ toward the plateau at $\Lambda^\star = \Lambda(t^\star)$. The longer time evolution is subject to fluctuations but doesn't alter significantly the picture. On the bottom panel, we confirm the predicted scaling $\Lambda^\star \sim \sqrt{p}$. We also show that the same scaling applies to the maximum of the global amplification factor, $\Lambda_\mr{max} = \mr{max}_t \, \Lambda(t)$, which occurs during the fitting regime.\\

In this section we illustrated with a particular example how the neurons of the network converge to a finite set of fixed points. The associated amplification factors are shown to diverge with the dataset size: $\lambda \sim \sqrt{p}$. At the network scale this effect is equivalent to a data compression of the same amplitude. In the next section we extend this discussion to other linear invariant datasets.

\section{Generalization and cylinder model}\label{sec:cylinder_model}

\paragraph{Compression mechanism}
The compression mechanism illustrated in the stripe model is expected to occur generically in linear invariant models. If the label function were to depend on $d_\parallel$ directions, all neuron vectors $\vl z$ would converge toward fixed points located in the informative subspace of dimension $d_\parallel$. Similar finite $p$ effects as in the stripe model would saturate the resolution of the informative subspace, so that the informative weights $\vl \omega_\parallel$ would be larger than the perpendicular weights $\vl \omega_\perp$ by an amplification factor $\lambda \sim \sqrt{p}$.

\paragraph{Advantage of feature regime}
As the NTK is blind to the existence of invariants in the data, the performance of the lazy regime should not depend on $d_\parallel$. Indeed following the results of \cite{paccolat2020isotropic}, the lazy regime learning curve follows an exponent $\beta_\mr{Lazy} = d / (3d -2)$ for simple boundaries separating labels (such as plane, spheres or cylinders), a result  conjectured to hold more generally for sufficiently  smooth boundaries. The correspondence between the lazy training and the SVC considered in \cite{paccolat2020isotropic} is discussed in \sref{sec:learning_curves}.

In \cite{paccolat2020isotropic}, it is also shown that for linear invariant models with $d_\parallel > 1$, a compression of the perpendicular space by a factor $\Lambda$, $\vl x_\perp \to \vl x_\perp / \Lambda$, improves the performance of the SVC by a factor $\Lambda^{-d_\perp / (3d- 2)}$, for a kernel of exponent $\xi = 1$. As discussed in \sref{sec:learning_curves}, because in the feature regime such a compression occurs with $\Lambda \sim \sqrt{p}$, we expect the learning curve exponent of the feature regime to be $\beta_\mr{Feature} = (d + \nicefrac{d_\perp}{2}) / (3d-2)$.

\paragraph{Cylinder model}

We test our predictions by considering a cylinder model in $d=3$. The data points are drawn from the standard normal distribution: $\vl x \sim \mc N(0, I_d)$, while the label function is a circle in the informative subspace of dimension $d_\parallel=2$, namely $y(\vl x) = y (\abs{\vl x_\parallel}) = +1$ if $\abs{\vl x_\parallel} > R$ and negative otherwise. For the numerical simulations we use $R = 1.1774$\footnote{This value is $\sqrt{2\log 2} \simeq 1.1774$. It is chosen so that the positive and negative labels are equiprobable.}. We learn this model following the gradient descent algorithm described in \sref{sec:learning_algorithm}. 

\begin{figure}[ht]
    \centering
    \includegraphics[scale=.34]{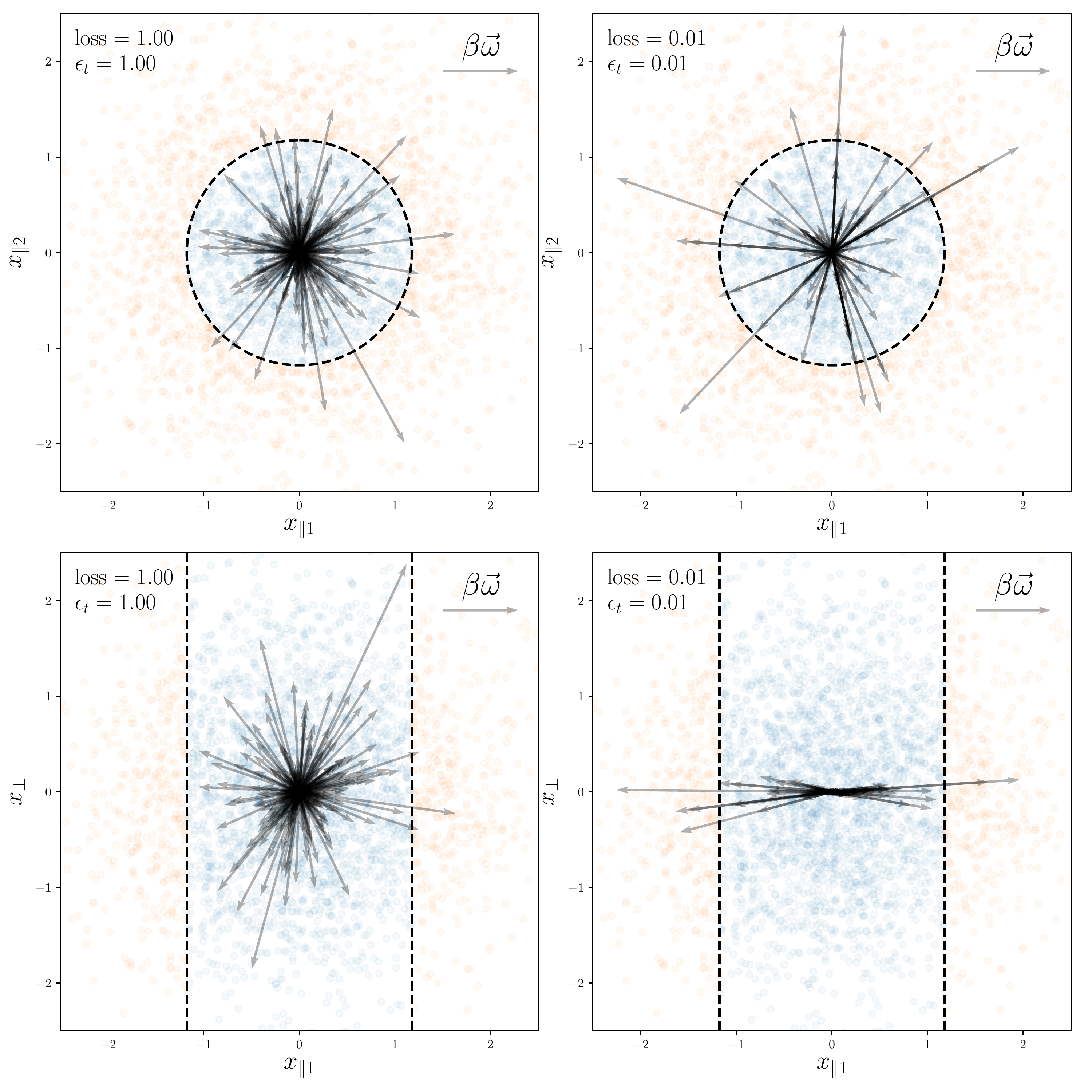}
    \caption{\label{fig:weight_orient_cylinder_fromdata} Representation of the amplification effect in the \textbf{cylinder model} with $d=3$ and $d_\parallel=2$. An instance of the labelled training set is shown in the background. The arrows represent the quantity $\beta \vl \omega$  -- properly rescaled to fit the figure -- for a random subset of neurons. We show the $x_\bot=0$ section (first row) and the $x_{\parallel2}=0$ section (second row) of data-space.
    The first column reports the weights distribution at initialization, the second column at the end of training. An animated version of the current figure can be found at \href{https://git.io/JJTS9}{git.io/JJTS9}.}
\end{figure}

\begin{figure}[ht]
    \centering
    \includegraphics[scale=.47]{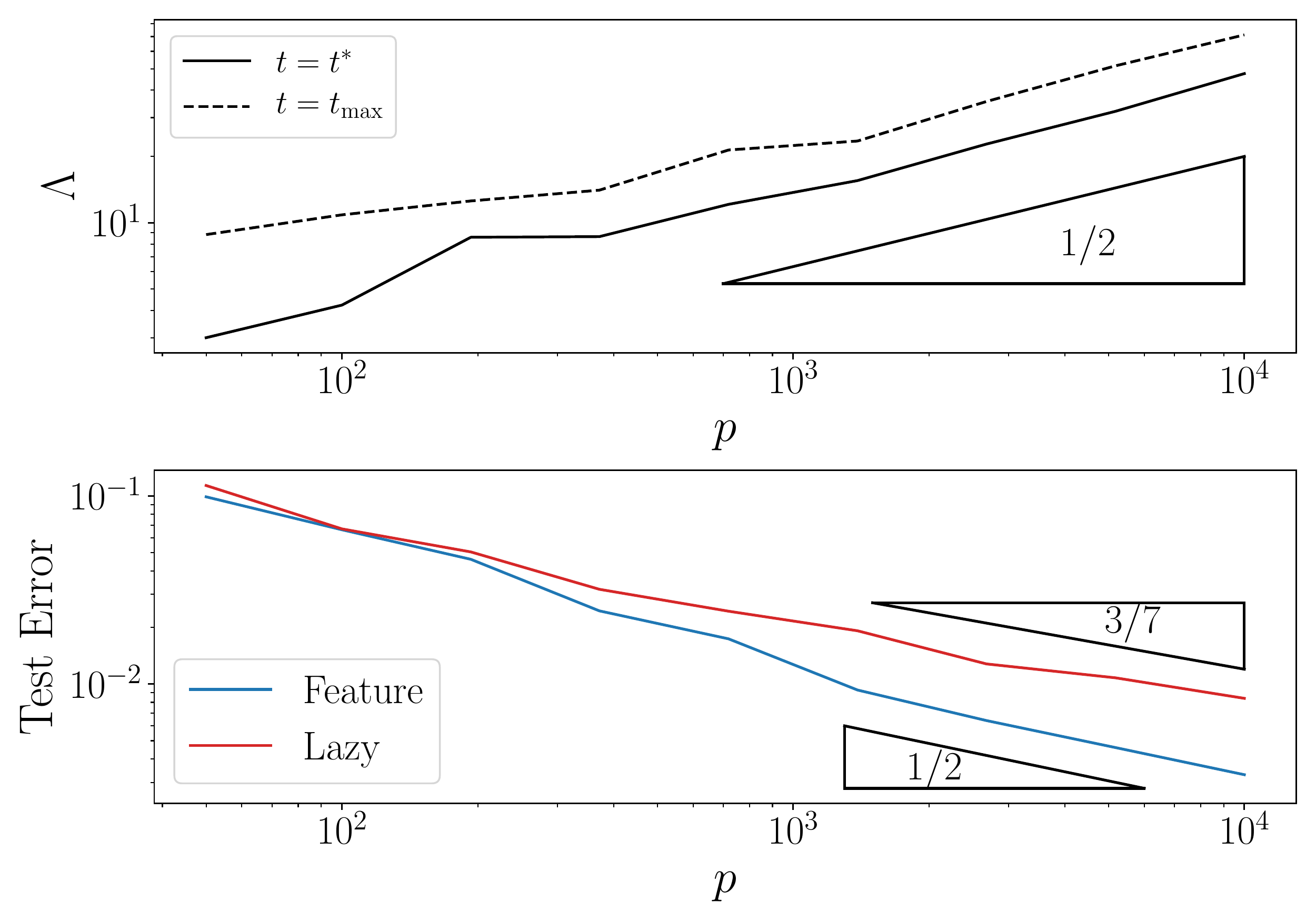}
    \caption{\label{fig:lambda_p_cylinder}  \underline{Top}: Global amplification factor vs the size of the training set $p$ for the \textbf{cylinder model} in dimension $d=3$. Both the amplification factor $\Lambda^\star$ computed at $t^\star$ and the maximal amplification factor $\Lambda_\mr{max}$ are displayed. The curves correspond to the median over 49 realizations of both the data distribution and the network initialization. The benchmark triangle illustrates the expected power law. \underline{Bottom}: Test error vs the training set size $p$ for the \textbf{cylinder model} in $d=3$. In the feature regime (blue line), the network scale is set to $\alpha = 10^{-6}$. In the lazy regime (red line), learning is performed with the frozen Gram matrix computed at initialization ($\alpha \to \infty$). The curves correspond to the median over 9 realizations of both the data distribution and the network initialization. The benchmark triangles illustrate the power law predictions.}
\end{figure}

The compression of the weight vectors $\vl \omega$ into the informative subspace displayed on \fref{fig:weight_orient_cylinder_fromdata} supports the previous general discussion. Also, we verify both the scaling of the amplification factor and the scaling of the learning curves on \fref{fig:lambda_p_cylinder}. As in the stripe model the time $t^\star$ is numerically defined as the time when the equality $y^\mu F(\vl x^\mu) > 1$ first holds for 10\% of the training set. On the top panel, both the global amplification factor at $t^\star$ and the maximal global amplification factor are shown to scale as $\Lambda^\star \sim \Lambda_\mr{max} \sim \sqrt{p}$. The advantage of the feature regime over the lazy regime is displayed on the bottom panel. In particular, the predicted learning curve exponents $\beta_\mr{Lazy} = d / (3d -2) = \nicefrac{3}{7}$ and $\beta_\mr{Feature} = (d + \nicefrac{d_\perp}{2}) / (3d-2) = \nicefrac{1}{2}$ are shown to be consistent with the numerical results.

\newpage
\section{Signatures of compression in the temporal evolution of the  NTK }\label{sec:shaping_kernel}
Previous empirical studies of compression of uninformative directions in data space in neural nets have focused on the  neural representations of the data layer by layer \cite{shwartz2017opening,recanatesi2019dimensionality}. Here instead we study 
how compression affects the evolution of the NTK as learning takes place, and show how this kernel becomes better suited for the considered task. We start from the stripe model and extend our analysis to a CNN trained on MNIST, and find striking similarities  between  the two cases.


\subsection{Neural Tangent Kernel Principal Components}
\label{sec:ntkpc}

\paragraph{General facts} The neural tangent kernel reads $\Theta(\vl x,\vl z) = \psi(\vl x) \cdot \psi(\vl z)$
where $\psi(\vl x)$ is a vector of $N$ components $\psi_W(\vl x) \vcentcolon = \partial_{W} f(\vl x)$ and $W$ is one of the $N$ parameters of the model. The kernel can be expressed in terms of its eigenvalues and eigenfunctions (Mercer's Theorem) $ \Theta(\vl x,\vl z) = \sum_\lambda \lambda \: \phi_\lambda(\vl x) \phi_\lambda(\vl z)$.
The functions $\phi_\lambda(\cdot)$ form an orthogonal basis on the space of functions, and satisfy the integral equation \cite{rasmussen_gaussian_2005} $\int \Theta(\vl x,\vl z)\phi_\lambda(\vl z)\rho(\vl z)d\vl z = \lambda \phi_\lambda(\vl x)$
where $\rho(\cdot)$ is the distribution of the data. In general, a kernel is expected to perform well  if the RKHS norm $\|y\|_\theta$ of the function $y(\vl x)$ being learnt is small \cite{scholkopf_kernel_1999}. It writes $\|y\|_\theta^2= \sum_\lambda \sfrac{\omega_\lambda^2}{\lambda}$ where $\omega_\lambda\vcentcolon = \int y(\vl x) \phi_\lambda(\vl x)\rho(\vl x) d\vl x$. Thus, a kernel performs better if the large coefficients $\omega_\lambda$ in the eigenbasis of the kernel correspond to large $\lambda$.  We will argue below that such a trend is enforced when the NTK evolves by compressing uninformative directions.

In practice, for a finite training set $\{\vl x^\mu\}_{\mu=1}^p$ of size $p$,   the Gram matrix $K$ is accessible empirically. It is defined as the $p \times p$ matrix of scalar products $ K_{\mu\nu} = \psi(\vl x^\mu) \cdot \psi(\vl x^\nu)$. Diagonalizing it corresponds to performing Kernel PCA \cite{scholkopf_kernel_1999}, which identifies the principal components in the feature representation $\psi(\vl x^\mu)$ of the data: $ K_{\mu\nu}  = \sum_{\tilde\lambda} \tilde\lambda \: \tilde\phi_\lambda(\vl x^\mu) \tilde\phi_\lambda(\vl x^\nu)$. One has $\tilde\lambda \rightarrow \lambda$ and $\tilde\phi_\lambda (\vl x^\nu)\rightarrow \phi_\lambda (\vl x^\nu)$ as $p\rightarrow\infty$ for a fixed $\lambda$. Thus the coefficients $\omega_\lambda$ can be estimated as $\tilde \omega_\lambda\vcentcolon = \tfrac{1}{p}\sum_{\mu=1...p} \tilde\phi_\lambda(\vl x^\mu)y(\vl x^\mu)$. In the following sections, we drop the tilde for ease of notation.


\paragraph{Effect of compression on the evolution of the NTK}
At initialization, for fully connected nets the NTK is isotropic, and its eigenvectors are spherical harmonics \cite{jacot2018neural}.
For a fixed dimension $d_\parallel$ of the informative space, as the overall dimension   $d$ grows, the value of a given spherical harmonics leads to vanishing information on the specific components $\vl x_\parallel$. As  a consequence, we expect that  even for large $\lambda$, $\phi_\lambda (\vl x)$  contains little information on the label $y(\vl x)$. It follows that the magnitude of the projected signal  $\omega_\lambda$ is small in that limit.

By contrast, after learning in the limit $\Lambda\sim \sqrt{p} \rightarrow\infty$, the output function looses its dependence on the orthogonal space $\vl x_\perp$. The NTK  can then generically be rewritten as:
\begin{equation}
\label{111}
    \Theta(\vl x,\vl z)=\Theta_1(\vl x_\parallel,\vl z_\parallel)+\Theta_2(\vl x_\parallel,\vl z_\parallel) \vl x_\perp \cdot \vl z_\perp
\end{equation}
where the second term comes from the derivative with respect to the first layer of weights (see \aref{app:ntk_decomposition}). 
For a Gaussian data density $\rho$ considered in this paper,
eigenvectors with non-vanishing eigenvalues   are then of two kind: $\phi_\lambda^1(\vl x_\parallel)$ -- the eigenvectors of $\Theta_1$ -- and $\phi_\lambda^2(\vl x_\parallel) \vl u\cdot \vl x_\perp$ where $\phi_\lambda^2(\vl x_\parallel)$ is an eigenvector of $\Theta_2$ and $\vl u$ any non-zero vectors. The null-space of the kernel then corresponds to all functions of the orthogonal space that are orthogonal to constant or linear functions. However for a finite $\Lambda$, we expect the associated eigenvalues to be small but different from zero.

Two qualitative predictions follow:
\begin{itemize}
    \item The eigenvectors $\phi_\lambda^1$ only depend on $\vl x_\parallel$ and are thus generically more informative on the label $y(\vl x_\parallel)$ than spherical harmonics. It is also true, but to a lesser extent, for the eigenvectors $\phi_\lambda^2(\vl x_\parallel) \vl u\cdot \vl x_\perp$. Indeed for Gaussian data, they can be considered as a function of $\vl x_\parallel$ times a random Gaussian noise.  Overall,  we thus expect that for large eigenvalues  the mutual information between $\phi_\lambda(\vl x)$ and $y(\vl x)$ to  increase during learning.
    
    \item As a consequence, the magnitude of  $\omega_\lambda$ associated to the top eigenvalues also tends to increase. We thus expect that the performance of  kernel learning using the NTK at the end of training to be superior to that using the NTK at initialization.
    
\end{itemize}

\begin{figure}[H]
    \centering
    \includegraphics[scale=.9]{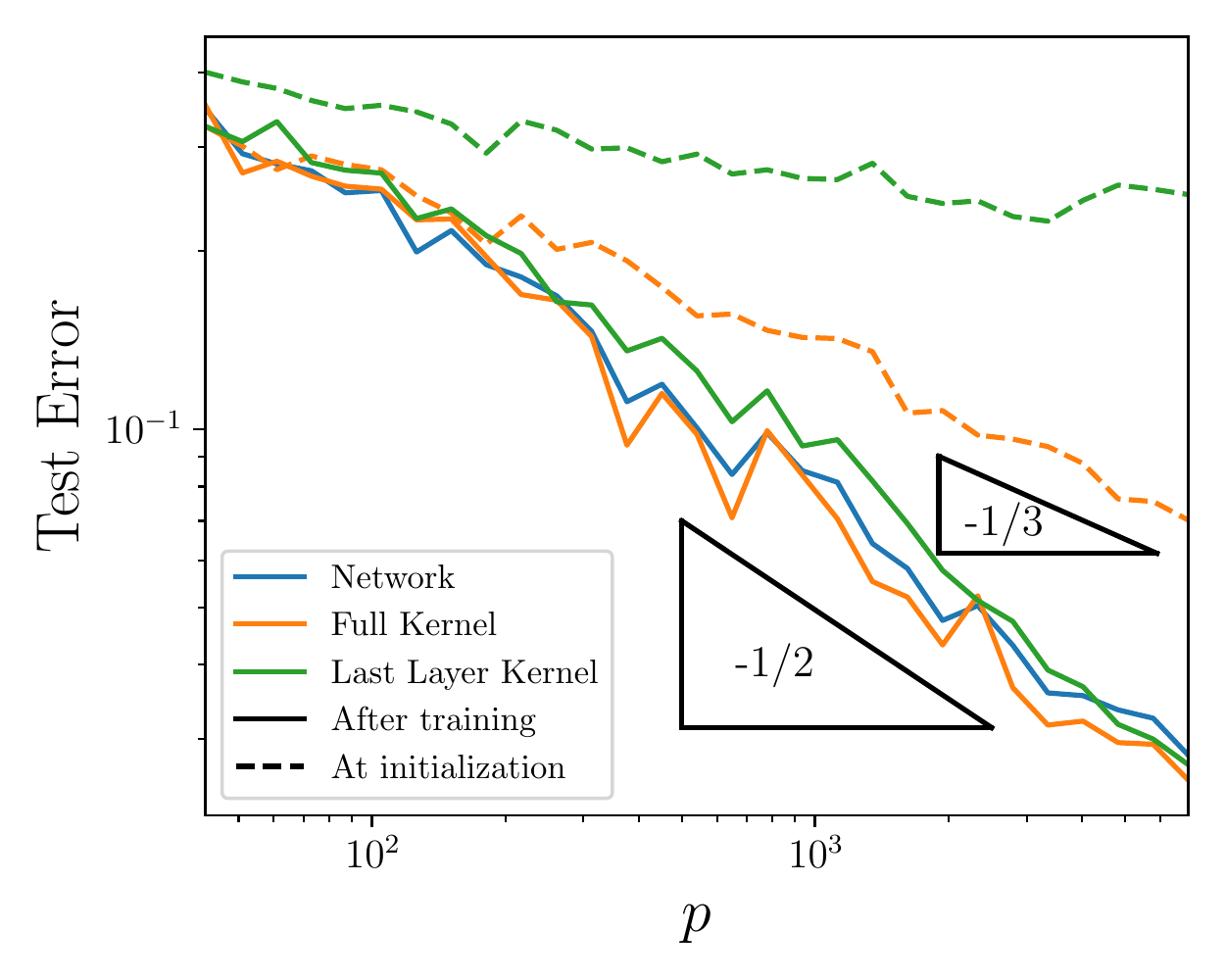}
    \caption{Performance of CNN trained to classify the parity of MNIST digits (binary class problem) as a function of the trainset size. The network is trained in the feature learning regime (blue line) using vanilla gradient descent with momentum, leading to $\beta_\mr{Feature}\approx 0.5$. Before and after training the full kernel (i.e. with respect to all the parameters) and the kernel of the last layer are computed. These four frozen kernels are then used in a gradient descent algorithm using an independent trainset of the same size. All the measures are done 5 times with different initialization seeds and averaged. For the full kernel at initialization (dashed orange) we find $\beta_\mr{Lazy} \approx 0.3$ and consequently $\beta_\mr{Lazy} <\beta_\mr{Feature}$.
    \label{fig:cnn}}
\end{figure}

\subsection{Empirical tests}

\paragraph{Performance of kernel methods based on the NTK} In Fig.\ref{fig:learning_curve} we test our prediction that kernel methods based on the NTK obtained at the end of training outperforms the NTK at initialization. We perform kernel learning using different data for the training set than those used to generate the NTK. We find that it is indeed the case: in fact, performance is found to be very similar to that of the neural net in the feature learning regime, except for the largest training set size where it even outperforms it. Note that this similarity is natural, since the features associated to the NTK contain the  the last hidden layer of neurons, which can represent the network output with the last layer of weights.

We test the generality of this result in Fig.\ref{fig:cnn} using a more modern CNN architecture on the MNIST data set. This architecture is inspired from MnasNet \cite{Tan_2019} with 16 convolutional layers. It distinguishes from MnasNet by the absence of batch-normalization. We again find that kernel methods based on the NTK at infinite time perform as well as the network in the feature learning regime, and even once again slightly better for the largest $p$. 

Finally, it is interesting to compare this analysis with the kernel whose features correspond to the last layer of hidden neurons  at the end of training. Training such a kernel simply corresponds to retraining the last layer of weights while fixing the activity of the last hidden neurons. Interestingly, this kernel performs well but generally less so than the network itself, as illustrated in 
Fig.\ref{fig:cnn}.



\begin{figure}[H]
    \centering
    \hspace{0.7cm}
    \vspace*{-.5cm}
    \includegraphics[scale=.5]{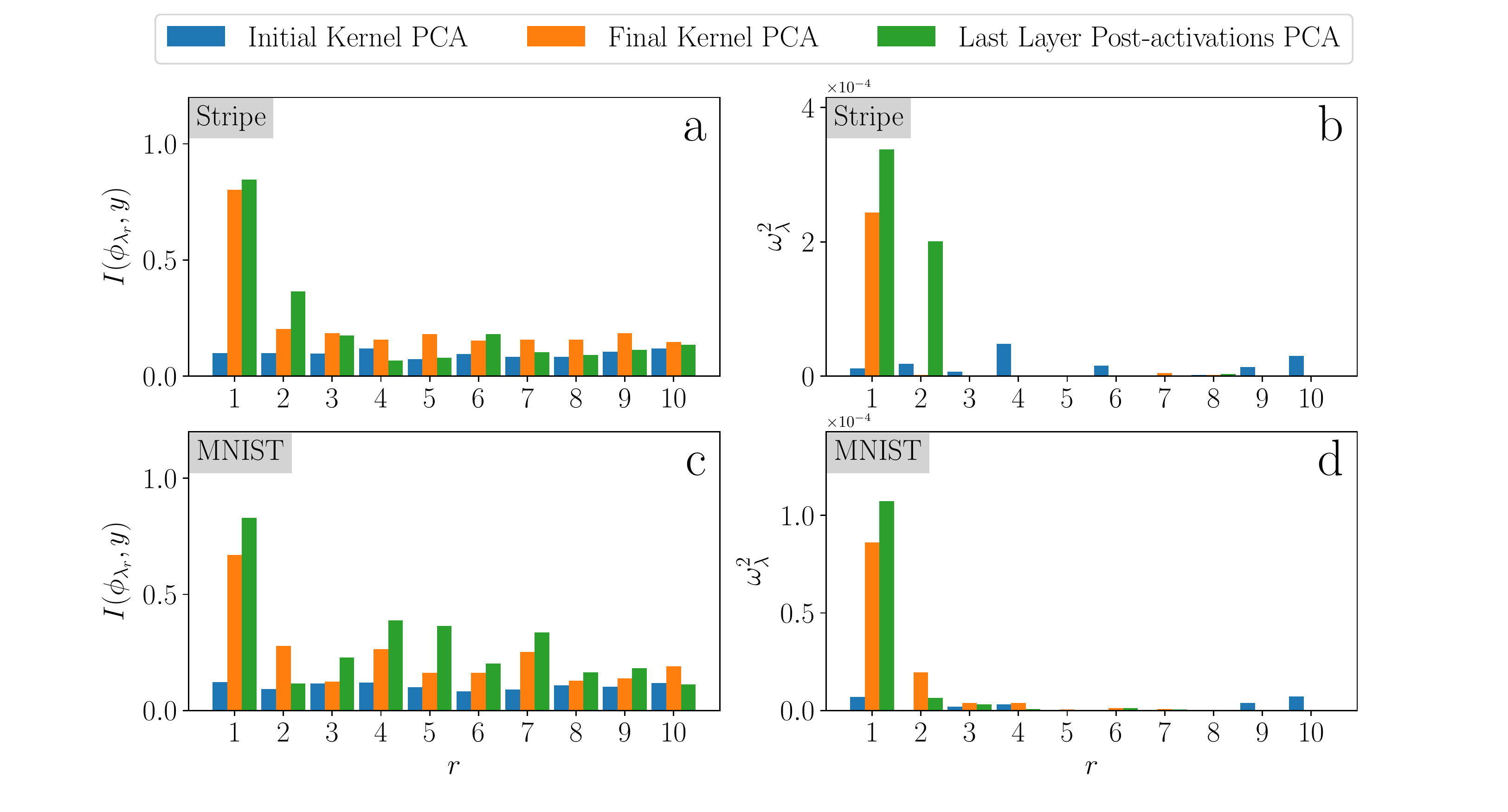}
    \caption{\label{fig:omegalambda_mi}
    Mutual Information $I(\phi_{\lambda_r}; y)$ between each of the first ten NTK eigenvectors and the output label for the \textbf{stripe model} (a) and \textbf{MNIST} (c), respectively. 
    The eigenvectors projection on the output labels $\omega_\lambda^2 = \frac{\langle \phi_{\lambda_r} | y\rangle^2}{p^2}$ is shown in panels (b) -- \textbf{stripe model} -- and (d) -- \textbf{MNIST}. We show in blue the results for the NTK at initialization, in orange for the NTK after training in the feature regime and in green for the principal components of last layer post-activations.}
\end{figure}

\paragraph{Kernel PCA {\it v.s.} labels (Information and projection)}
We now confirm that such improved performance of the NTK corresponds to the top kernel principal components becoming more informative on the task. As we argued in Section \ref{sec:ntkpc}, we expected this to be the case, in the presence of compression. 
Specifically, we consider the $r$ largest eigenvalues $\lambda_\text{max}, \dots \lambda_r$ of the NTK Gram Matrix and their corresponding eigenvectors. We first compute the mutual information between a given eigenvector magnitude and the label $I(\phi_{\lambda_r}; y)$ -- for details on the estimator see Appendix \ref{sec:mi_appendix}. 
This mutual information is small and essentially independent of $r$ in the range studied for the NTK at initialization; both for the stripe model (Fig.\ref{fig:omegalambda_mi}.a) and MNIST (Fig.\ref{fig:omegalambda_mi}.c). However, at the end of learning, mutual information has greatly improved in both cases, a fact that holds true for the NTK and for the kernel obtained from the last layer of hidden neurons. 

As expected, the magnitude of the  projection of each of the first $r$ eigenvectors onto the output labels $\omega_\lambda^2 = \langle \phi_{\lambda_r} | y\rangle^2 / p^2$ also greatly improves during learning.
This effect is striking both for the stripe model  (Fig.\ref{fig:omegalambda_mi}.b) and for MNIST (Fig.\ref{fig:omegalambda_mi}.d). At initialization, that projection does not show a significant trend with rank within the first 10 eigenvectors. Yet after learning, most of the projection occurs along the first mode of the NTK alone, with the second mode also showing a sizable projection for MNIST.

\begin{figure}[H]
    \centering
    \hspace*{-1cm}
    \includegraphics[scale=.533]{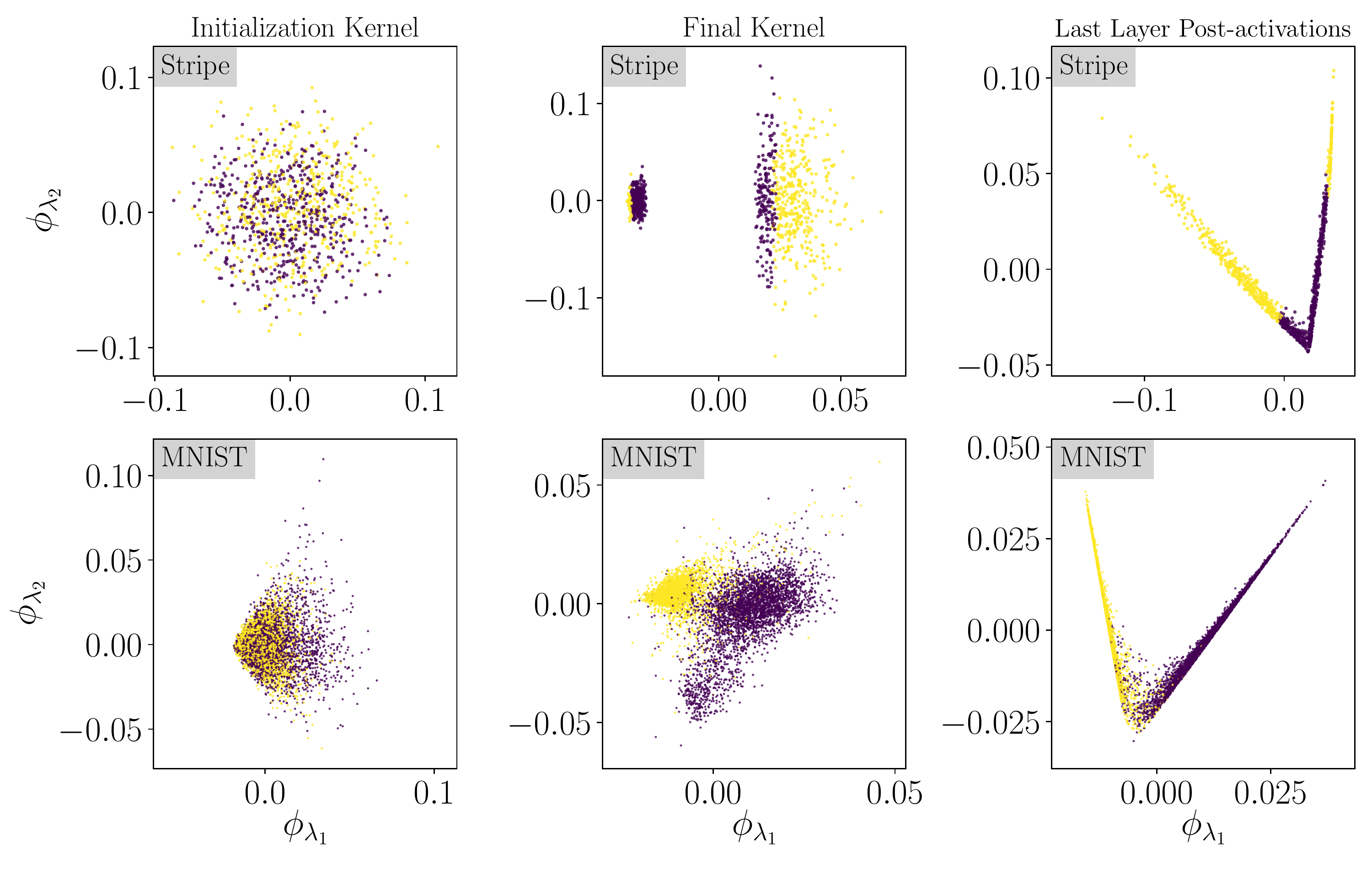}
    \caption{Scatter plot of the first two NTK eigenvectors -- $\phi_{\lambda_1}$ and $\phi_{\lambda_2}$ -- for the \textbf{stripe model} (first row) and \textbf{MNIST} (second row).  Colors map class labels. Eigenvectors are computed for the NTK at initialization (first column) and after training (second column). The last column refers to the last layer post-activation principal components. These results are consistent with \fref{fig:omegalambda_mi}: (1) Before learning, the eigenvectors are not correlated to the labels, while they are after learning. (2) For the stripe model, only the first eigenvector of the final kernel contains information on the labels, as expected from panel b) of \fref{fig:omegalambda_mi}. (3) Two informative eigenvectors are necessary to linearly separate the stripe data as illustrated on the top-right panel as well as on panel b) of \fref{fig:omegalambda_mi}. The associated unidimensional representation is expected from the effective data compression for the stripe model where $d_\parallel=1$. (4) For MNIST, the first two eigenvectors of the final kernel are not sufficient to classify completely the data as expected from panel d) of \fref{fig:omegalambda_mi}, but still suggest a compression along the uninformative directions. This last point is also motivated by the approximate unidimensional collapse observed in the bottom-right panel.
    }
    \label{fig:features_scatter}
\end{figure}

Overall, the similarities of these plots between  MNIST and the stripe model support that compression is indeed a key effect characterizing learning for MNIST as well. To study further these similarities, we focus on the first two eigenvectors and plot data points (different labels appear as different colors) in the $(\phi_{\lambda_1}(\vl x),\phi_{\lambda_2}(\vl x))$ plane as shown in \fref{fig:features_scatter}. As expected, these eigenvectors at initialization have essentially no information on the output label -- the scatter plot looks like Gaussian noise both for the stripe model and MNIST (left column). By contrast, after learning data of different classes appear as well separated clouds of points in that plane (central column). Strikingly, performing the same analysis for the kernel obtained from the last layer of hidden neurons shows that data organize into a smaller manifold, which is approximately one-dimensional (right column). It is expected in the stripe model, since for $\Lambda\rightarrow\infty$ the hidden neurons activity can only depend on a single variable $x_1$. It is interesting that a similar dimension-reduction appears so clearly in MNIST as well, suggesting the importance of a nearly-one dimensional manifold in the representation of the last hidden layer. We have checked that such a one-dimensional structure is not apparent in the effective dimension of this representation  \footnote{ Computing the effective dimension (based on the scaling of the distance between points in terms of the number of points \cite{Facco_2017}) of that representation leads to $d_{\text{eff}}\approx 6$, possibly coming from the finite width of the nearly one-dimensional manifold apparent in  \fref{fig:features_scatter}, bottom right.}.

\section{Conclusion}\label{sec:conclusion}

We have shown that in the presence of $d_\perp$ uninformative dimensions of the input, the weights of a one-hidden layer neural network become orthogonal to them. 
For a vanishingly small initialization of the weights and vanilla gradient descent, this effect is limited by the sample noise of the training set,  and its magnitude is of order $\Lambda\sim \sqrt{p}$. For simple geometries of the boundaries separating labels, this effect increases the exponent $\beta$ characterizing learning curves with respect to the lazy training regime (in which the neuron orientation is nearly frozen). This increase depends on both  $d_\perp$ and $d$.   Both for the stripe and cylindrical model, the observed exponents $\beta$ are consistent with this prediction, supporting that for these models at least the main advantage of the feature learning regime is to compress invariant directions.

Next we have argued that such a compression shapes the evolution of the neural tangent kernel during learning, so that its principal components become more informative and display a larger projection on the label, effectively reducing the RKHS norm of the function being learnt. As a consequence, using gradient descent with the frozen NTK at the end of training leads to much better performance than at initialization, and we observe that it even outperforms the neural net in the feature learning regime. The analysis underlines that kernel PCA on the NTK is a valuable tool to characterize the compression of invariants.  Overall we find striking similarities between a one-hidden layer FC network trained on the stripe model and a deep CNN trained on MNIST, supporting that compression is central to the performance of the latter as well.  

One challenge for the future is to classify  which conditions on the data can guarantee such an improvement of the NTK during learning -- a question directly connected to the relative performance of lazy training {\it v.s.} feature learning, which appears to depend on the architecture for real data \cite{geiger2019}.

A second challenge is the development of quantitative models for the compression of other symmetries in the data, including the invariance of the label toward smooth deformations that characterize images. Is this compression  ultimately responsible for the success of deep learning in beating the curse of dimensionality ? Answering this question  presumably requires to focus on more modern architectures, in particular deep CNNs.

\subsection*{Acknowledgments}
We acknowledge G. Biroli, M. Gabrie, D. Kopitkov, S. Spigler, Y. Rouzaire and all members of the PCSL group for discussions. This work was partially supported by the grant from the Simons Foundation (\#454953 Matthieu Wyart). M.W. thanks the Swiss National Science Foundation for support under Grant No.~200021-165509.

\clearpage
\printbibliography
\clearpage
\appendix

\section{Rotation invariance}\label{app:rotation_invariance_2}
In this appendix we prove that if we rotate the input of the network it doesn't affect its performance. \\

\textbf{Lemma}: For a group $G$ and a $G$-invariant function $f$, the gradient of $f$ is $G$-equivariant: $$\nabla f(D(g)\vl x) = D(g)^{-T} \nabla f(\vl x) \quad \forall g\in G \,\mr{and} \, \forall\vl x,$$
where $D$ is the representation of $G$ acting on the space of inputs $\vl x$ and $A^{-T}$ denotes the inverse transpose of the matrix $A$.

\textbf{Proof} The derivative of $f$ in the direction $u$ evaluated in $D(g) x$ is given by
\begin{align}
  \vl  u \cdot \nabla f(D(g) \vl x) &= \lim_{h\to0} \frac{f(D(g)\vl x + h\vl u) - f(D(g)\vl x)}{h} \\
                             &= \lim_{h\to0} \frac{f(\vl x + h D(g)^{-1}\vl u) - f(\vl x)}{h} \\
                             &= (D(g)^{-1}\vl u) \cdot \nabla f(\vl x) = u \cdot (D(g)^{-T} \nabla f(\vl x)).
\end{align}
Since this formula holds for any direction $\vl u$, it proves the lemma.\\

In the context of a neural network, if the loss function of a neural network satisfies $\mathcal{L}(D_w(g)\vl w, D_x(g)\vl x) = \mathcal{L}(\vl w,\vl x)$ with $D_w$ orthogonal, it is easy to see that the lemma applied to the loss reads $\nabla_w \mathcal{L}(D_w(g)\vl w, D_x(g)\vl x) = D_w(g) \nabla_w \mathcal{L}(\vl w,\vl x)$. Here $\vl w$ refers to the weights of the network, whose dynamics is given by $\vl{\dot{w}}(t) = - \sum_\mu \nabla_w \mathcal{L}(\vl w, \vl{x}_\mu)$, where $\mu$ is the training set index. If we act with $G$ on $\vl w$ and on the training set, the derivative $\dot{\vl{w}}$ is transformed in the same way as $\vl w$. A network initialised to $D_w(g) \vl{w}_0$ instead of $\vl{w}_0$ and trained on $\{ D_x(g) \vl{x}_\mu \}_\mu$ instead of $\{ \vl{x}_\mu \}_\mu$ during a time $t$ will thus have its weights equal to $D_w(g) \vl{w}(t)$ instead of $\vl{w}(t)$.

In particular, this discussion holds for a network starting with a fully-connected layer: in this case $G$ is the orthogonal group, $D_x$ is the orthogonal matrix and $D_w$ is acting on the first weights with an orthogonal matrix and leaves the rest of the weights invariant.

In case of an initialisation distribution of the weights that satisfies $\rho(D_w(g) w) = \rho(w)$, the expected performance (averaged over the initialisations) will be independent of the global orientation of the inputs.

\section{\texorpdfstring{$\alpha$}{alpha} scan in the stripe model}\label{app:alpha_scan}

\begin{figure}[ht]
    \centerline{\includegraphics[width=0.97\textwidth]{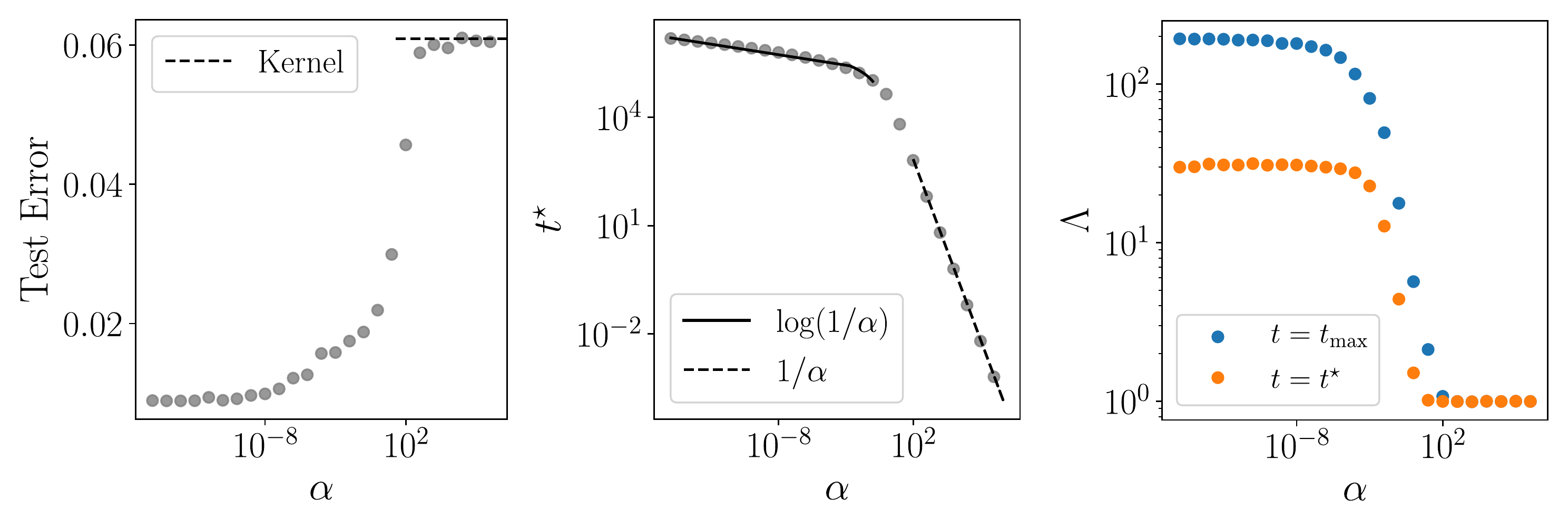}}
    \caption{\label{fig:scan_alpha}  On each plot, the dots are obtained by averaging the gradient descent results over 5 different data realizations and network initialization. \underline{Left}: Test error vs the network scale $\alpha$. The horizontal dashed line correspond to the test error of the frozen initial kernel dynamics, also averaged over 5 realizations. \underline{Center}: Characteristic time $t^\star$ vs the network scale $\alpha$. \underline{Right}: Global amplification factor vs the network scale $\alpha$. Both the amplification factor at $t^\star$ and the maximal amplification factor are represented.}
\end{figure}

We illustrate the transition from the feature regime to the lazy regime by considering the single-stripe model in dimension $d=10$ with a training set of size $p=1000$. We vary the network scale from $\alpha=10^{-16}$ to $\alpha=10^8$ (see \fref{fig:scan_alpha}). In the limit $\alpha \to \infty$, the test error converges to the one obtained by running the kernel dynamics with the NTK frozen at initialization, the characteristic time scales as $t^\star \sim 1/\alpha$ as expected from \cite{geiger2019} and the global amplification factor equals one. In the opposite limit, $\alpha \to 0$, the test error converges to a plateau better than the lazy regime performance, the characteristic time grows logarithmically as discussed in \sref{sec:timescales} and the global amplification factor reaches a plateau. 

\section{Stripe model dynamics}\label{app:stripe_analytics}

In this section, we give additional details to the computation carried in \sref{sec:compressing_regime}. We consider the large $p$ limit of the system \eqref{eq:neuron_dynamics1}, where it is well approximated by the central-limit theorem. For $t \ll t^\star$, the dynamics of each neuron is governed by the system
\begin{align*}\label{eq:ap_ODE}
\dot{\omega}_1 &=  \frac{\beta}{\tau} \: \Big(\mu_1 + \frac{\sigma_1}{\sqrt{p}} N_1 \Big) \\
\dot{\vl \omega}_\perp &=  \frac{\beta}{\tau} \: \Big(\vl \mu_\perp + \frac{\sigma_\perp}{\sqrt{p}} \vl N_\perp \Big) \\
\dot{b} &= \frac{\beta}{\tau} \: \Big(\mu_b + \frac{\sigma_b}{\sqrt{p}} N_b \Big)
\end{align*}
up to $\mc O(p^{-1})$ corrections. The last layer weight is obtained from the constant of motion $\beta^2 - \abs{\vl \omega}^2 - b^2 = \mr{const}$. We compute the averages $\mu_1$, $\vl \mu_\perp$ and $\mu_b$ in \aref{app:computation_averages} and discuss the asymptotic solution in the limit $p\to\infty$ in \aref{app:infinite_p}. The finite $p$ corrections and the associated standard deviations $\sigma_1$, $\sigma_\perp$ and $\sigma_b$ are considered in \aref{app:finite_p}.

\subsection{Computation of the averages}\label{app:computation_averages}
We compute the averages $\mu_1$, $\vl \mu_\perp$ and $\mu_b$ for data distributed according to the standard normal distribution: $\rho(\vl x) = \rho(\abs{\vl x}) = (2\pi)^{-\nicefrac{d}{2}} \exp\left(-\abs{\vl x}^2/2\right)$. For the bias and the informative weights, we get
\begin{align*}
    \mu_1 &= \int \mr{d} \vl x \rho(\vl x) x_1 y(x_1) \Theta\left[\frac{\vl \omega \cdot \vl x}{\sqrt{d}} + b \right] \\
    &= \int \mr{d}x_1 \rho(x_1) x_1 y(x_1) \frac{1}{2}\left[1 + \mr{erf}\left(\frac{b\sqrt{d} + \omega_1 x_1}{\sqrt{2} \omega_\perp}\right)\right] \\
    \mu_b &= \sqrt{d} \int \mr{d} \vl x \rho(\vl x) y(x_1) \Theta\left[\frac{\vl \omega \cdot \vl x}{\sqrt{d}} + b \right] \\
    &= \sqrt{d} \int \mr{d}x_1 \rho(x_1) y(x_1) \frac{1}{2}\left[1 + \mr{erf}\left(\frac{b\sqrt{d} + \omega_1 x_1}{\sqrt{2} \omega_\perp}\right)\right].
\end{align*}
For the perpendicular weights, we treat each components independently, so that for $i>1$:
\begin{align*}
    \mu_i &= \int \mr{d} \vl x \rho(\vl x) x_i y(x_1) \Theta\left[\frac{\vl \omega \cdot \vl x}{\sqrt{d}} + b \right] \\
    &=\frac{\mr{sgn}(\omega_i)}{\sqrt{2\pi}}\int \mr{d}x_1 \rho(x_1) y(x_1) \int \mr{d}x_\perp \rho(x_\perp) \mr{e}^{-\frac{(b\sqrt{d} + \omega_1 x_1 + \tl\omega_\perp x_\perp)^2}{2 \omega_i^2}} \\
    &= \frac{1}{\sqrt{2\pi}} \frac{\omega_i}{\omega} \mr{e}^{-\frac{db^2}{2\omega^2}} \int \mr{d}x_1 \rho(x_1) y\left(\frac{\omega_\perp}{\omega}x_1 - \frac{\sqrt{d} b \omega_1}{\omega^2}\right) \\
\end{align*}
where we used the notation $\tl \omega_\perp = \sqrt{\omega_\perp^2 - \omega_i^2}$. Using the definition of $\lambda$ and $\zeta_1$, one thus recovers the system \eqref{eq:neuron_dynamics2}.

\subsection{Infinite \texorpdfstring{$p$}{trainset size}}\label{app:infinite_p}

\paragraph{Expectation values in the limit $|\lambda| \to \infty$}
In the limit $\lambda \to \pm \infty$, the function $g_\lambda$ becomes a Heaviside function whose direction depends on the sign of $\lambda$: $g_\lambda(x) \xrightarrow{\lambda \to \pm \infty} \Theta(\pm x)$. Consequently, the remaining integrals over the $x_1$ distribution in \aref{app:computation_averages} simplifies:
\begin{align*}
    \< y(x_1) \, x_1 \, g_\lambda(x_1 - \zeta_1)\>_{x_1} &\xrightarrow{\lambda \to \pm \infty} \< y(x_1) \, x_1 \, \Theta\left(\pm(x_1 - \zeta_1)\right)\>_{x_1} = C_1^\pm(\zeta_1) \\
    \< y(x_1) \, g_\lambda(x_1 - \zeta_1)\>_{x_1} &\xrightarrow{\lambda \to \pm \infty} \< y(x_1) \, \Theta\left(\pm(x_1 - \zeta_1)\right)\>_{x_1} = C_b^\pm(\zeta_1) \\
    \< y\left(\frac{\omega_\perp}{\omega}x_1 + \frac{\omega_1^2}{\omega^2}\zeta_1\right)\>_{x_1} &\xrightarrow{\lambda \to \pm \infty} y(\zeta_1) = C_\perp(\zeta_1) \\
\end{align*}

\paragraph{Asymptotic solutions}

We assume that the neuron vector $\vl z$ is set constant and equal to $\vl z^\star = (z^\star, \vl 0)$. The dynamics of $\omega_1$, $\beta$ and $b$ thus no longer depend on the perpendicular weights. In the asymptotic regime, the sign of $\omega_1 \beta$ is given by the sign of the constant $C_1^\pm(z^\star)$. In particular, using the constant of motion and the definition $b = -\omega_1 z^\star / \sqrt{d}$, we get $\beta = \mr{sign}\left(C_1^\pm(z^\star)\right) \sqrt{1 + \nicefrac{z^{\star 2}}{d}} \, \omega_1$, where we neglected the order one value of the constant of motion. Finally, one finds that the informative weights diverge as
\begin{equation}
    \omega_1 \sim \mr{e}^{t/\tau^\star}, \:\: \mr{with} \: \tau^\star = \frac{\tau}{|C_1^\pm(z^\star)|\sqrt{1 + \nicefrac{z^{\star 2}}{d}}}.
\end{equation}
Inserting the above relations into the perpendicular weights dynamics yields
\begin{equation}
    \dot{\vl \omega}_\perp = \mr{sign}\left[C_1^\pm(z^\star) \omega_1 y(z^\star)\right] \frac{\sqrt{1 + \nicefrac{z^{\star 2}}{d}}}{2 \pi \tau} \mr{e}^{-\frac{z^{\star 2}}{2}} \, \vl \omega_\perp.
\end{equation}
Hence, if $\mr{sign}\left[C_1^\pm(z^\star) \omega_1 y(z^\star)\right] = -1$, the perpendicular weights all vanish exponentially. However, if $\mr{sign}\left[C_1^\pm(z^\star) \omega_1 y(z^\star)\right] = +1$, they all diverge exponentially with a time constant
\begin{equation*}
    \tau^\star_\perp = \frac{2 \pi \tau}{\sqrt{1 + \nicefrac{z^{\star 2}}{d}}} \, \mr{e}^{\frac{z^{\star 2}}{2}}
\end{equation*}
which still leads to a diverging amplification factor if $\tau^\star_\perp > \tau^\star$.

\subsection{Finite \texorpdfstring{$p$}{trainset size}}\label{app:finite_p}
We assess the finite $p$ corrections  of the asymptotic solutions given in \aref{app:infinite_p}. Since the bias and the informative weights are divergent, they are not sensitive to finite $p$ corrections. However, for the perpendicular weights, it is essential to compute the standard deviations. Since the expectations have been computed previously, it is sufficient to look at the second non-central moments. For simplicity, we directly consider the limit $|\lambda| \to \infty$, so that for $i>1$:
\begin{align*}
    \mu_i^2+\sigma_i^2 &= \int \mr{d} \vl x \rho(\vl x) x_i^2 \Theta\left[\frac{\vl \omega \cdot \vl x}{\sqrt{d}} + b \right] \xrightarrow{\lambda \to \pm \infty} \int \mr{d} x_1 \rho(x_1) \Theta\left[\pm (x_1 - z^\star) \right] = D^\pm_\perp(\zeta_1).
\end{align*}
For each perpendicular direction, a random variable of variance one quantifies the discrepancy between the average $\mu_i$ and the exact sum over the dataset. Its value depends on the location of the ReLU hyperplane. In particular, once the considered neuron has reached its fixed point $\vl z^\star$, all random variables can be arranged into the constant perpendicular vector $\vl N_\perp(z^\star)$.

\section{NTK decomposition and eigenfunctions}\label{app:ntk_decomposition}

In section \ref{sec:ntkpc} we argued that, for the setting considered in this paper, the NTK can be decomposed as 
\begin{equation}
\label{theta_appendix}
    \Theta(\vl x,\vl z)=\Theta_1(\vl x_\parallel,\vl z_\parallel)+\Theta_2(\vl x_\parallel,\vl z_\parallel) \vl x_\perp \cdot \vl z_\perp.
\end{equation}
In this appendix, we look at this decomposition more in details and derive the eigenfunctions functional form.

\paragraph{NTK decomposition} Recall the architecture considered in this paper, 
\begin{equation*}
f(\vl x) = \frac{1}{h} \sum_{n=1}^h \beta_n \: \sigma\left(\frac{\vl \omega_n \cdot  \vl x}{\sqrt{d}} + b_n\right).
\end{equation*}
For this architecture, the NTK reads
\begin{equation*}
    \Theta(\vl x, \vl z) = \frac{1}{h^2}\sum_{n=1}^h \Bigg[ \sigma\left(\frac{\vl \omega_n \cdot \vl x}{\sqrt{d}} + b_n\right) \sigma\left(\frac{\vl \omega_n \cdot \vl z}{\sqrt{d}} + b_n\right) + \beta_n^2 \sigma^\prime\left(\frac{\vl \omega_n \cdot \vl x}{\sqrt{d}} + b_n\right) \sigma^\prime\left(\frac{\vl \omega_n \cdot \vl z}{\sqrt{d}} + b_n\right) \left(1 + \frac{\vl x \cdot \vl z}{d}\right) \Bigg].
\end{equation*}
If the input space has only $d_\parallel$ informative directions, after feature learning ($\Lambda \to \infty$), the output function will only depend on $\vl x_\parallel$. This is because $\vl \omega_n \cdot \vl x \to \vl \omega_{n,\parallel} \cdot \vl x_\parallel$ and the NTK can be rewritten as 
\begin{multline*}
    \Theta(\vl x, \vl z) = \frac{1}{h^2}\sum_{n=1}^h \Bigg[ \sigma\left(\frac{\vl \omega_{n,\parallel} \cdot \vl x_\parallel}{\sqrt{d}} + b_n\right) \sigma\left(\frac{\vl \omega_{n,\parallel} \cdot \vl z_\parallel}{\sqrt{d}} + b_n\right) \\ 
    + \beta_n^2 \sigma^\prime\left(\frac{\vl \omega_{n,\parallel} \cdot \vl x_\parallel}{\sqrt{d}} + b_n\right) \sigma^\prime\left(\frac{\vl \omega_{n,\parallel} \cdot \vl z_\parallel}{\sqrt{d}} + b_n\right) \left(1 + \frac{\vl x_\parallel \cdot \vl z_\parallel}{d} + \frac{\vl x_\bot \cdot \vl z_\bot}{d}\right) \Bigg],
\end{multline*}
where one can readily identify $\Theta_1(x_\parallel, \vl z_\parallel)$ and $\Theta_2(x_\parallel, \vl z_\parallel)$.

\paragraph{NTK eigenfunctions} Eigenfunctions satisfy the integral equation 
\begin{equation*}
    \int \Theta(\vl x,\vl z)\phi_\lambda(\vl z)\rho(\vl z)d\vl z = \lambda \phi_\lambda(\vl x),
\end{equation*}
where $\rho(\cdot)$ is the distribution of the data. We assume here that $\rho(\vl x) = \rho_\parallel(\vl x_\parallel) \rho_\perp(\vl x_\perp) = \rho_\parallel(\vl x_\parallel) \Pi_i \rho_\perp(x_{\perp, i})$ with zero mean and the same variance in all directions. If we plug in the decomposition (\ref{theta_appendix}), we notice that eigenvectors are of two kinds, they are either eigenvectors of $\Theta_1(\vl x_\parallel,\vl z_\parallel)$ or of $\Theta_2(\vl x_\parallel,\vl z_\parallel) \vl x_\perp \cdot \vl z_\perp$ -- i.e. they give zero when the other operator acts on them. The ones coming from $\Theta_1$ are solutions of
\begin{equation*}
    \int \Theta_1(\vl x_\parallel,\vl z_\parallel)\phi^1_\lambda(\vl z)\rho(\vl z)d\vl z = \lambda \phi_\lambda^1(\vl x).
\end{equation*}
Given that the l.h.s. only depends on $\vl x_\parallel$, we have $\phi_\lambda^1(\vl x) = \phi_\lambda^1(\vl x_\parallel)$. Integrating out $\vl z_\bot$ we get
\begin{equation*}
    \int \Theta_1(\vl x_\parallel,\vl z_\parallel)\phi^1_\lambda(\vl z_\parallel)\rho(\vl z_\parallel)d\vl z_\parallel = \lambda \phi_\lambda^1(\vl x_\parallel).
\end{equation*}
The second kind of eigenvectors satisfy
\begin{equation*}
    \int \Theta_2(\vl x_\parallel,\vl z_\parallel){\vl x_\bot \cdot \vl z_\bot}\phi^2_\lambda(\vl z)\rho(\vl z)d\vl z = \lambda \phi_\lambda^2(\vl x).
\end{equation*}
Notice that $\vl x_\bot$ can be moved out of the integral. Consequently, eigenfunctions can only linearly depend on the perpendicular component -- i.e. $\phi_\lambda^2(\vl x) = \phi_\lambda^2(\vl x_\parallel )\, \vl u \cdot \vl x_\bot$. The integral equation reads
\begin{equation*}
    \left(\int \Theta_2(\vl x_\parallel,\vl z_\parallel)\phi^2_\lambda(\vl z_\parallel)\rho(\vl z_\parallel)d\vl z_\parallel \right)  (\vl u \cdot \vl x_\bot) \int z_\bot^2 \rho(z_\bot)d z_\bot= \lambda \phi_\lambda^2(\vl x_\parallel )\, \vl u \cdot \vl x_\bot,
\end{equation*}
where $\vl u$ can be any non-zero vector.
To back what we stated previously -- i.e. that eigenvectors are either of the two kinds -- we show that no other eigenvector, different from linear combinations of $\phi_\lambda^1$ and $\phi_\lambda^2$, exists. Assume there exists $\phi^*_\lambda(\vl x) \neq a \phi^1_{\lambda_1}(\vl x) + b \phi^2_{\lambda_2}(\vl x)$, this would solve
\begin{align*}
    \int \left[\Theta_1(\vl x_\parallel,\vl z_\parallel)+\Theta_2(\vl x_\parallel,\vl z_\parallel) \vl x_\perp \cdot \vl z_\perp\right] \phi^*_\lambda(\vl z)\rho(\vl z)d\vl z &= \lambda \phi^*_\lambda(\vl x) \\
    \int \Theta_1(\vl x_\parallel,\vl z_\parallel) \phi^*_\lambda(\vl z)\rho(\vl z)d\vl z +\int \Theta_2(\vl x_\parallel,\vl z_\parallel) \vl x_\perp \cdot \vl z_\perp \phi^*_\lambda(\vl z)\rho(\vl z)d\vl z &= \lambda \phi^*_\lambda(\vl x) \\
    \lambda_1 \phi^1_{\lambda_1}(\vl x) + \lambda_2 \phi^2_{\lambda_2}(\vl x) &= \lambda \phi^*_\lambda(\vl x),
\end{align*}
resulting in a contradiction.

\section{Mutual Information Estimator}
\label{sec:mi_appendix}

We propose a mutual information estimator $\widehat{I}(\bx;y)$ that exploits the information we know about the binary labels distribution $P(y)$:
\begin{equation*}
    P(y=+) = P(y=-) = \frac{1}{2}.
\end{equation*}
The variable $\bx$ is continuous and can live in high dimension.
We define
\begin{equation*}
    q_+ = \frac{P(\bx|y=+)}{2P(\bx)} = P(y=+|\bx), \qquad q_- = \frac{P(\bx|y=-)}{2P(\bx)} = P(y=-|\bx).
\end{equation*}

We recall the definition of differential entropy for continuous variables,
\begin{equation*}
    H(\vl x) = - \int d^d\bx P(\bx) \log P(\bx)
\end{equation*}

Given that the mutual information can be expressed $I(\bx;y) = H(\bx) - H(\bx|y)$, we compute the conditional entropy knowing $P(y)$ as\footnote{All the logarithms of this section are computed in base 2.}
\begin{align*}
    H(\bx|y) &= -\frac{1}{2} \int d^d\bx \: \pplus \log \pplus -\frac{1}{2} \int d^d\bx \: \pminus \log \pminus  \\
             &= -\int d^d\bx \: \p q_+ \log (2\p q_+) - \int d^d\bx \: \p q_- \log (2\p q_-) \\
             &= -\int d^d\bx \: \p (q_+ + q_-) \log (2\p) - \int d^d\bx \: \p \big[ q_+ \log (q_+) + q_- \log (q_-)\big]\\
             &= H(\bx) - 1 - \E_{\bx} \big[ q_+ \log (q_+) + q_- \log (q_-)\big]. 
\end{align*}

Finally, the mutual information is given by
\begin{align*}
    I(\bx;y) &= H(\bx) - H(\bx|y) \\
             &= 1 + \E_{\bx} \big[ q_+ \log (q_+) + q_- \log (q_-)\big].
\end{align*}

We find the following estimator 
\begin{align*}
    \widehat{I}(\bx;y) &= 1 + \frac{1}{p} \sum_{i=1}^p\:  \widehat{q_+}(\vl x_i) \log \widehat{q_+}(\vl x_i) + \widehat{q_-}(\vl x_i) \log \widehat{q_-}(\vl x_i) \\
                       &= 1 - \frac{1}{p} \sum_{i=1}^p\: h_2(\widehat{q_+}(\vl x_i)),
\end{align*}
where $h_2(\cdot)$ is the binary entropy function\footnote{$h_2(x) = -x\log x - (1-x)\log (1-x)$ .}.

We notice that we can rewrite 
\begin{align*}
    P(y|\bx) &= \frac{P(\bx|y)}{2\sum_y P(\bx|y)P(y)} \\
             &= \frac{P(\bx|y)}{P(\bx|y=+) + P(\bx|y=-)},
\end{align*}
hence the MI estimation reduces to estimating $P(\bx|y)$. 

At this stage we propose the following approximation: suppose that $P(\bx|y)$ is uniform in the ball containing the $k$ nearest neighbors of $\bx$ which are labelled $y$, i.e.\footnote{The estimation depends on the value of $k$ which is omitted to simplify the notation. For the estimations in this paper we use $k=5$.}
\begin{align*}
    P(\bx| y) &\sim r_{y}^{-d}(\vl x_i).
\end{align*}
The estimation of $\widehat{q_+}$ finally reduces to
\begin{equation}
\label{eq:qhat}
    \widehat{q_+}(\vl x_i) = \frac{r_+^{-d}(\vl x_i)}{r_+^{-d}(\vl x_i) + r_-^{-d}(\vl x_i)} \\
                       = \frac{1}{1 + \left(\frac{r_+(\vl x_i)}{r_-(\vl x_i)}\right)^d}
\end{equation}
\begin{equation*}
        \widehat{q_-}(\vl x_i) = 1 - \widehat{q_+}(\vl x_i).
\end{equation*}

We tested the estimator on different datasets and identified two main flaws:
\begin{itemize}
    \item For large $d$, the estimator gets affected by the \textit{curse of dimensionality}, distances between data-points become all similar to each other. As a result, the $\widehat{q_+}$ estimator gets biased towards $\nicefrac{1}{2}$.
    \item If $\vl x$ lives on a manifold of dimension lower than the one of the embedding space, the use of $d$ in \eref{eq:qhat} -- instead of the effective local dimension around $\vl x_i$ -- biases the estimator towards its extrema.
\end{itemize}
Considering we employ the estimator only in $d=1$, we skip the discussion on the possible ways to correct these flaws.



\end{document}